\documentclass[preprint,5pt]{elsarticle}

\usepackage{amssymb}
\usepackage{amsmath}
\allowdisplaybreaks[4]
\usepackage{hyperref}
\usepackage{pifont}
\usepackage{xcolor}
\usepackage{dsfont}
\usepackage{booktabs}  
\usepackage{multirow}  
\usepackage{graphicx}
\usepackage{array}
\usepackage{tabularx}
\usepackage{bbm}
\usepackage{subcaption}
\usepackage{float}
\usepackage{adjustbox}  
\usepackage{caption}   
\usepackage{siunitx}
\usepackage{makecell}
\usepackage{lineno}
\usepackage{placeins}

\newcommand{\best}[1]{$\textbf{#1}$}
\newcommand{\second}[1]{\underline{#1}}

\newcommand{\bestx}[1]{\textcolor{red}{\textbf{#1}}}
\newcommand{\secondx}[1]{\textcolor{blue}{\underline{#1}}}

\journal{Knowledge-Based Systems}
\begin{document}

\begin{frontmatter}



\title{FM-LLM: A Frequency-Enhanced Mixture-of-Experts Framework for Adapting LLMs to Time Series Forecasting} 


\author[a]{Rentao Gu} 
\author[a]{Yihang Ding}
\author[b]{Junjie Li}
\author[b]{Yi Ding}
\author[a]{Weijing Sang}
\author[b]{Xiaoli Huo}
\author[b]{Xin Qin}
\author[a]{Yuefeng Ji}

\affiliation[a]{organization={Beijing University of Posts and Telecommunications},
	city={Beijing},
	postcode={100876},
	country={China}}

\affiliation[b]{organization={China Telecom Research Institute},
	city={Beijing},
	postcode={100032},
	country={China}}

\begin{abstract}
Recent advances in Large Language Models (LLMs) have spurred cross-modal solutions for time-series forecasting. However, existing methods rely heavily on textual prompts for modality alignment---introducing nontrivial computational overhead and failing to leverage the rich spectral dynamics inherent in time-series data. To enable prompt-free, frequency-aware adaptation of frozen LLMs, we propose FM-LLM (\textbf{F}requency-Enhanced \textbf{M}ixture-of-Experts for adapting \textbf{LLM}s to Time Series Forecasting), an autoregressive framework grounded in the principle of constrained asymmetric coupling. A Fourier Analysis Network (FAN)-based spectral token aligner injects structured harmonic representations directly into the frozen LLM with numerical compatibility. Concurrently, an asymmetric Mixture-of-Experts (MoE) decoder enforces explicit role separation: shared experts equipped with lightweight FAN layers reconstruct the global periodic backbone, while routed experts---restricted exclusively to standard feed-forward networks (FFNs)---specialize in modeling non-periodic residual dynamics. A time-frequency hybrid loss function jointly optimizes temporal accuracy and spectral consistency, effectively mitigating error accumulation during long-horizon autoregressive rollouts. Evaluated across eleven public benchmarks, FM-LLM achieves state-of-the-art performance on 59 out of 78 evaluation metrics. Compared to the strongest autoregressive LLM-based baseline, it delivers average improvements of 5.3\% in Mean Squared Error (MSE) and 5.6\% in Mean Absolute Error (MAE), with maximum gains reaching 8.0\% for MSE and 8.4\% for MAE. FM-LLM also demonstrates robust transferability, maintaining superior performance in 10\% few-shot and zero-shot forecasting scenarios. By purposefully wiring spectral alignment and decoding---distinct from prompt-dependent alignment strategies or uniform spectral processing approaches---FM-LLM establishes an accurate, efficient, and interpretable paradigm for next-generation time-series LLMs.

\end{abstract}
\end{frontmatter}







\section{Introduction}
\label{sec1}
Time-series forecasting is widely applied in domains such as energy systems~\cite{10850726}, traffic management~\cite{zhao2019t}, network planning~\cite{fiandrino2024aichronolens}, and climate modeling~\cite{Kochkov2024NeuralGCM}, where accurate long-horizon predictions are essential for reliable resource allocation, anomaly detection, and strategic planning. As data volumes and real-time decision requirements continue to grow, the demand for robust and interpretable forecasting solutions has never been more pressing. Recent advances in deep learning—ranging from MLPs and CNNs to Transformer-based architectures—have significantly improved forecasting accuracy by modeling nonlinear temporal dependencies. More recently, pre-trained Large Language Models (LLMs) have been explored for time-series forecasting due to their strong transferability under limited supervision (few-shot/zero-shot)~\cite{NEURIPS2020_1457c0d6} and powerful context modeling ability, which can condition predictions on long historical contexts and auxiliary cues; such context-based conditioning has also been leveraged in recent LLM-based forecasters~\cite{liu2024autotimes}. Moreover, their text-based interface makes it convenient to incorporate side information when available, and can potentially be extended to multimodal settings~\cite{yin2024survey}.

Despite these advantages, adapting LLMs to time-series forecasting remains challenging. First, time series are continuous-valued signals whereas LLMs operate on discrete tokens, making modality alignment non-trivial and potentially compromising numerical fidelity. Second, real-world time series exhibit multi-scale periodic and non-periodic dynamics that are not naturally captured by text-pretrained inductive biases~\cite{LIU2025113478}, and purely time-domain prompting/tokenization may miss critical spectral structures. To mitigate these issues, existing LLM-based forecasting methods typically rely on prompts and lightweight encoders to provide the LLM with periodicity-related information and map continuous signals into token-like representations.

For instance, GPT4TS~\cite{zhou2023one} fine-tunes a partially frozen GPT-2; Time-LLM~\cite{jin2023time} reformulates time windows as text-like sequences; AutoTimes~\cite{liu2024autotimes} embeds time segments via MLPs and augments them with index prompts. PRADA~\cite{LIU2025113478} first applies season--trend decomposition to the input series, encoding the decomposed features as learnable prompts. 

Despite their differences, all these methods \textit{invariably adopt a patch-based strategy} to align time-series inputs with the language modality, and \textit{employ shallow linear decoders}---typically one or two layers---to map LLM outputs back to the time-series domain. While effective to some extent, these design choices raise a key concern: they may oversimplify the complex temporal and spectral structure inherent in multivariate time-series data.

This motivates us to revisit the fundamental design assumptions and ask two central questions:

\begin{itemize}
    \item \textbf{Can we extract richer and more structured representations from time-series patches, beyond treating them as flat tokens, to better inform LLMs?}
    \item \textbf{In multivariate forecasting, where variables exhibit diverse temporal and spectral characteristics, is a single linear decoder sufficient to model such heterogeneous patterns effectively?}
\end{itemize}

For the first question, time-series data often embed complex and multi-scale periodicities, which are difficult to fully capture through token-level representations in the time domain alone. In fact, many existing forecasting models have demonstrated the effectiveness of frequency-domain analysis in modeling such temporal structures—for instance, FEDformer leverages Fourier attention to highlight dominant components~\cite{zhou2022fedformer}, while TimesNet selects informative frequency bases to enhance prediction~\cite{wu2023timesnet}. These observations motivate us to use spectral decomposition not merely as an auxiliary feature, but as a structured modality translator that aligns continuous-valued patches with the discrete token-wise computation of LLMs.

For the second question, multivariate time-series data often exhibit diverse temporal patterns across different variables, with varying periodicity, phase shifts, and noise levels. To address such heterogeneity at the same granularity as LLM token processing, we draw inspiration from the Mixture-of-Experts (MoE) architecture and perform patch-token-wise conditional routing. This enables different experts to specialize in distinct temporal regimes while keeping the overall computation efficient.

Adapting a frozen LLM for time-series forecasting introduces unique and stringent requirements beyond generic modality alignment. The fundamental challenge lies in efficiently injecting high-density frequency information and disentangling its superimposed entanglement with temporal dynamics—all while strictly avoiding prompt engineering, which typically incurs prohibitive inference latency through sequence expansion. To address these needs, we propose \textbf{FM-LLM} (\textbf{F}requency-Enhanced \textbf{M}ixture-of-Experts for adapting \textbf{LLM}s to time-series forecasting), a framework that couples explicit spectral token alignment with constrained conditional decoding in a purposefully asymmetric manner. 

On the encoding side, a Fourier Analysis Network~\cite{dong2024fan} (FAN)-based spectral token aligner maps each patch into a structured harmonic representation, establishing an efficient, prompt-free information pathway directly into the frozen LLM. On the decoding side, we enforce a design philosophy of explicit role separation: only the always-on shared (Fourier) expert incorporates lightweight FAN layers to reconstruct the global periodic backbone, whereas routed experts are intentionally restricted to standard FFNs to specialize in irregular, non-periodic residuals. This constrained asymmetric coupling not only prevents spectral bias from contaminating residual specialization but also effectively accommodates the diverse temporal patterns across variables in multivariate sequences through token-wise conditional computation. Extensive experiments on long-horizon forecasting benchmarks demonstrate that \textbf{FM-LLM} achieves state-of-the-art performance, obtaining the best results on 59 out of 78 evaluation metrics.

\paragraph{Our main contributions are summarized as follows:}

\begin{itemize}
    \item \textbf{Spectrum-Time Decoupling for Frozen LLM Adaptation.} We propose a \textit{prompt-free} spectral token aligner that replaces traditional text-based prompts with high-density harmonic representations. This design establishes an efficient information pathway into frozen LLMs, bypassing the inference latency overhead caused by sequence expansion.

    \item \textbf{Constrained Asymmetric Decoding Strategy.} We introduce a heterogeneous MoE decoder with explicit structural constraints to resolve the \textit{superimposed entanglement} of temporal dynamics. By restricting spectral modeling to shared experts and standard FFNs to routed experts, we prevent spectral bias from contaminating residual specialization, effectively disentangling periodic backbones from non-stationary residuals.

    \item \textbf{Holistic Time-Frequency Optimization.} We design a hybrid loss function that jointly supervises temporal accuracy and spectral consistency. This dual-domain objective reinforces the structural decoupling of the decoder and significantly mitigates error accumulation in long-horizon autoregressive forecasting.
\end{itemize}

\section{Related Work}

\subsection{Multivariate Time-Series Forecasting}
Traditional time-series methods, such as AutoRegressive Integrated Moving Average (ARIMA)~\cite{box1970distribution}, exponential smoothing~\cite{gardner1985exponential}, and spectral analysis~\cite{percival1993spectral}, have long been foundational in analyzing temporal data. These statistical models play an important role in revealing trends, seasonality, and structural patterns. However, their linear assumptions and stationarity requirements often limit their capacity to model complex non-linear dependencies and long-range temporal dynamics in real-world data.

In recent years, deep learning approaches have become dominant in time-series forecasting due to their high capacity for non-linear representation learning. Common paradigms include CNN-based, GNN-based, Transformer-based, MLP-based, and more recently, LLM-based and KAN-based models. Alongside network architectures, auxiliary techniques such as sequence decomposition, spectral filtering, and Fourier analysis have been shown to further improve model precision by injecting structural inductive biases.

CNN-based methods such as MICN~\cite{wang2023micn} and TimesNet~\cite{wu2023timesnet} are notable. MICN adopts a multiscale downsampling convolution combined with globally equivariant convolutions, enabling efficient linear-complexity modeling that separates trend, periodicity, and seasonal components. TimesNet adaptively reshapes 1D sequences into 2D tensors based on multiple inferred periods and captures intra- and inter-period variations using lightweight multiscale Inception modules.

GNN-based methods, such as MSGNet~\cite{cai2024msgnet} and PANDA~\cite{10889936}, explicitly model inter-series dependencies via graph structures. MSGNet captures multi-scale correlations by combining frequency decomposition with adaptive graph convolutions. PANDA integrates patching into graph networks, utilizing a dual time-frequency alignment strategy to optimize spatio-temporal representations.

MLP-based methods like DLinear~\cite{Zeng_Chen_Zhang_Xu_2023} and TimeMixer~\cite{wang2023timemixer} demonstrate that strong forecasting performance can be achieved with simple, efficient architectures. DLinear employs linear projection and decomposition, outperforming more complex models with minimal structure. TimeMixer proposes an MLP-based multiscale mixing framework tailored to different sampling rates.

Transformer-based models like PatchTST~\cite{Yuqietal-2023-PatchTST} and iTransformer~\cite{liu2023itransformer} utilize patch-based encoding and token-level attention to model temporal dependencies. PatchTST segments time series into overlapping patches and embeds them into latent space, while iTransformer treats each univariate series as a token and models inter-series dependencies using self-attention and intra-series dynamics via FFNs.

LLM-based forecasting models, including GPT4TS~\cite{zhou2023one}, AutoTimes~\cite{liu2024autotimes}, Time-LLM~\cite{jin2023time}, CVC~\cite{LI2025113957} and EV-STLLM~\cite{fan2026evstllm}, adapt pretrained language models to time-series tasks. GPT4TS fine-tunes LayerNorm and position encoding of the first six GPT2 layers to leverage few-shot capabilities. Time-LLM aligns numerical tokens with language models via prefix-style prompts. AutoTimes encodes sliced segments with MLPs in an autoregressive fashion to mimic LLM training dynamics. CVC addresses the modality gap by employing a dual-branch architecture with contrastive correction to align time-series and textual features. EV-STLLM integrates multi-frequency decomposition with a partially frozen graph attention module to adapt LLMs for complex spatio-temporal forecasting.

Recent innovations include heterogeneous MLPs like TimeKAN~\cite{huang2025timekan}, which introduces a Decomposition-Learning-Mixing framework using frequency decomposition, multi-order Kolmogorov-Arnold networks (KANs)~\cite{liu2024kan}, and frequency mixing to learn multifrequency temporal patterns effectively.

\subsection{Fourier Analysis Network}
Fourier-based neural networks have evolved from early shallow architectures using sinusoidal activations to deep networks integrating spectral transforms. Initial works like Fourier Neural Networks~\cite{gallant1988there} employed cosine basis functions as activations to approximate periodic functions compactly. These models showed promise in control systems and signal approximation~\cite{zuo2005fourier, zuo2008approximation}, but were limited in depth and flexibility.

Subsequent studies identified limitations in ReLU-based networks for learning periodic signals. Liu et al~\cite{ziyin2020neural} demonstrate that standard neural networks struggle to learn periodic functions due to the lack of inductive bias, and propose a sinusoid-based activation function to address this limitation. Empirical studies by Uteuliyeva~\cite{uteuliyeva2020fourier} further confirmed the benefits of Fourier activations in shallow models.

Recent efforts have introduced Fourier representations into deeper architectures. SIREN~\cite{sitzmann2020implicit} builds MLPs with sine activations, showing breakthroughs in signal reconstruction and PDE solving. More relevantly, FAN~\cite{dong2024fan} integrates Fourier expansions into each layer using sin/cos projections of intermediate features, achieving strong extrapolation capabilities across language, vision, and time-series tasks. Unlike earlier domain-specific Fourier networks, FAN is designed as a general-purpose module and has proven especially effective in modeling periodic dynamics.

\subsection{Mixture of Experts (MoE)}
Recent Large pretrained models have increasingly adopted the Mixture-of-Experts (MoE) paradigm. MoE divides model capacity into specialized components 'experts', each responsible for specific sub-tasks. Only some experts respond to each individual input, enabling efficient inference while retaining model diversity. This scalable framework enables growth in model capacity without proportional increases in compute cost.

MoE models such as Mixtral-8x7B~\cite{jiang2024mixtral}, Switch Transformer~\cite{fedus2022switch}and DeepSeek-V3~\cite{liu2024deepseek} have demonstrated strong performance across tasks. However, MoE architectures face challenges such as training instability and expert load imbalance. DeepSeek addresses these with an auxiliary-loss-free load balancing mechanism that adjusts routing bias per expert. To prevent extreme imbalance within individual sequences, it introduces a sequence-wise balance loss to ensure intra-sequence fairness.

This aligns well with autoregressive forecasting where each token (or variable) may represent different patterns. In multivariate settings, even within the same time series, heterogeneous modes exist. MoE enables the model to learn such patterns and dynamically assign them to specialized experts. Our experiments show that different MoE experts indeed capture distinct temporal dynamics, improving overall forecasting performance.

\subsection{Loss Functions in Time and Frequency Domains}

Existing research has explored loss function design in both the time and frequency domains to improve training effectiveness and prediction accuracy in time-series forecasting.

Time-domain losses aim to directly minimize prediction error at each time step. Standard objectives like Mean Squared Error (MSE) and Mean Absolute Error (MAE) treat all horizons equally, which may lead to degraded performance for long-range forecasting. To address this, Signal Decay-based Loss (SDL)~\cite{xue2024card} introduces a horizon-aware weighting scheme that emphasizes short-term predictions while suppressing noisy long-term targets. Similarly, the Loss Shaping Constraint (LSC)~\cite{hounie2024loss} applies step-specific error thresholds to control cumulative errors during multi-step forecasting.

Frequency-domain losses leverage spectral representations to better capture periodic structures in the data. FreDF: Learning to Forecast in the Frequency Domain~\cite{wang2025fredf} highlights that direct multi-step loss may misalign with the true data distribution due to autocorrelation. It proposes computing errors in the frequency domain to decouple temporal dependencies and enhance training stability. These frequency-based losses are often combined with MSE to balance time-domain accuracy and spectral consistency.

Inspired by both paradigms, we design a hybrid loss that combines SDL in the time domain and spectrum-based loss in the frequency domain. This formulation reinforces short-horizon prediction fidelity while promoting frequency-aware representation learning, which consistently improves performance across benchmarks.


\section{Methodology}


\subsection{Problem Definition}

Given a historical multivariate time series $\textbf{x}_{1:T} = \{ x_1,x_2 ,\ ...,x_T\} \in \mathbb R^{C\times T}$, the objective of multivariate time-series forecasting is to predict a future sequence $\mathrm{{\textbf{x}}}_{T+1:T+F} = \{x_1,x_2 ,...,x_F\} \in \mathbb R^{C\times F}$, where $ T$ and $F$ denote the lengths of the look-back window and forecast horizon, respectively, and $C$ represents the number of variables to be predicted. The proposed FM-LLM is, an adaptation of large language models tailored for multivariate time-series forecasting. The goal of this task is to train a frozen large language model (LLM)-based forecaster to predict the next $F$ time steps using the past $T$ steps, formulated as:
\begin{equation}
f: ( \mathrm{\textbf{x}}_{1:T}) \mapsto \hat{\mathrm{x}}_{L+1:L+F}.\tag{1}
\end{equation}

\subsection{Modality Alignment}
\begin{figure*}
    \centering
    \includegraphics[width=1\linewidth]{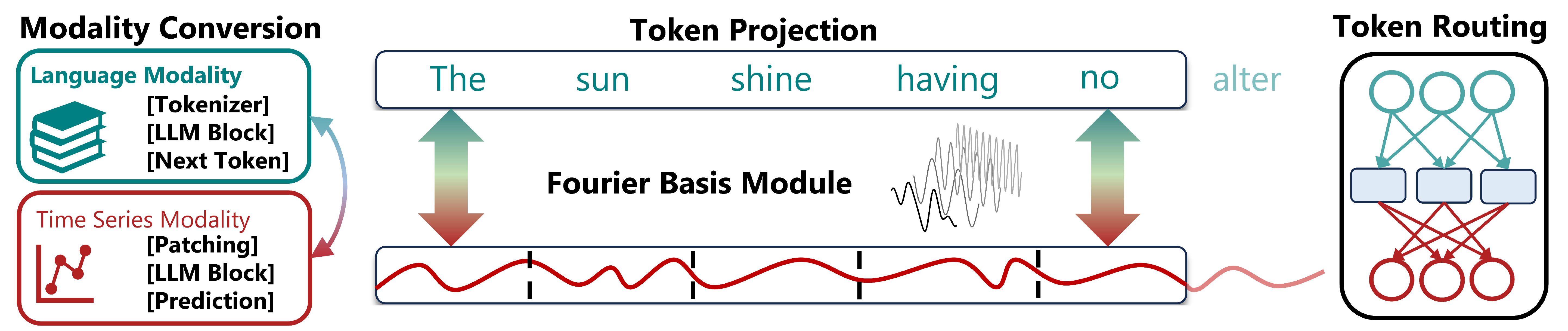}
    \caption{An example to illustrate how FM-LLM adpts large language models for time series forecasting.}
    \label{fig:1}
\end{figure*}
\subsubsection{Tokenization}
We adopt the concept of patching as shown in Figure \ref{fig:1} to convert the time series into a token tensor and incorporate the idea of channel-wise independence\cite{Yuqietal-2023-PatchTST}.
Let the uni-variate input time series be denoted as $\mathrm{\textbf{x}}_{1:T}\in\mathbb{R}^T$, where $P$ represents the token length and $S$ denotes the stride between two consecutive, non-overlapping tokens. This patching process segments the time series into short subsequences, resulting in $\mathrm{\textbf{x}}_{p}\in\mathbb{R}^{N\times P}$, where $N$ is the total number of tokens, computed as $N=\left\lfloor \frac{T-P}{S} +1\right\rfloor$. In this work the token length and stride are set to $P=S$. By adopting such a tokenization strategy, the number of tokens can be reduced from $T$ to approximately $T/{S}$, which leads to a substantial reduction in both memory usage and computational complexity of the attention mechanism. This benefit is particularly significant when deploying large-scale Transformer models such as LLMs, because it can effectively reduce inference time and resource consumption.

\subsubsection{Token Embedding}
We follow the autoregressive prediction framework\cite{liu2024autotimes} and repurpose autoregressive LLMs as time series forecasters as depicted in Figure \ref{fig:1}. Let ${x}_t^i\in \mathbb{R}$ denote the value of the $i$-$\mathrm{th}$ variable at time step $t$. The $k$-$\mathrm{th}$ segment of length P is defined as follows:
\begin{equation}
    {P}_k=\{x_{(k-1)P+1},...,x_{kP} \}\in \mathbb{R}^P,k=1,2,...,N.\tag{2}
\end{equation}

To leverage the transferability of LLMs, their parameters are kept frozen.To align the discrete language tokens with the time-series tokens, a token embedding function is defined as:$\mathrm{Tokenembedding}(\cdot)$ :$\mathbb{R}^P \mapsto \mathbb{R}^H$ which projects each time-series token into the latent space of the LLM. The embedded tokens are represented as:
\begin{equation}
TE_k=\mathrm{Tokenembedding}(P_k), k=1,2,..,N,\tag{3}
\end{equation}

where $H$ denotes the hidden dimension of the LLM. The token embedding process is described in the following section.
\subsubsection{Fourier Embedding Module}

\begin{figure*}[!t]
    \centering
    \includegraphics[width=1\linewidth]{figure_2.eps}

    \caption{The input time series is first divided into tokens, which are then passed into the encoder. In the lower-left part of the figure, each token is decomposed into sine and cosine basis functions, as indicated by the red lines, while the blue lines connect to activation functions. The outputs of the LLM are routed through a Mixture-of-Experts (MoE) module. In the lower-right part of the figure, the right-side diagram illustrates that the Fourier experts represent the LLM outputs as weighted combinations of sine and cosine functions, with 'FC' denoting fully-connected layers. $\mathrm{P}_1-\mathrm{P}_4$ and $\mathrm{P}_2-\mathrm{P}_5$ indicate the temporal ordering of input tokens and output tokens. During training, we compute the loss only for the final output token.}
    \label{fig:merged}
\end{figure*}

Fourier theory states that any practically meaningful periodic function can be represented as a Fourier series—that is, a weighted sum of complex exponential functions that are harmonically related and share the same period as the function being represented. Because time series often contain complex periodic patterns, introducing the Fourier series is necessary. The Fourier Analysis Network~\cite{dong2024fan} aims to construct an implicit periodic model within the data, expressed as follows:
\[
\mathrm{FAN}(x) = \phi_L \circ\phi_{L-1}\circ \cdots \circ \phi_1 \circ x \tag{7}
\]
where
\[
\phi (x) =\begin{cases}
[cos(W_p^lx)||sin(W_p^lx)||\sigma(B_{\bar p}^l + W_{\bar p}^lx)], & \text{if }  l<L,\\
B^L + W^Lx, & \text{if } l=L,
 \end{cases}
 \tag{8}
\]
where $W_p \in \mathbb{R}^{d_x \times d_p}$, $W_{\bar{p}} \in \mathbb{R}^{d_x \times d_{\bar{p}}}$, and $B_{\bar{p}} \in \mathbb{R}^{d_{\bar{p}}}$ are learnable parameters (with the hyperparameters $d_p$ and $d_{\bar{p}}$ indicating the first dimension of $W_p$ and $W_{\bar{p}}$, respectively), the layer output $\phi(x) \in \mathbb{R}^{2d_p + d_{\bar{p}}}$, and $\sigma$ denotes the activation function. In this paper, we use only a single-layer Fourier analysis network (FAN).
Figure \ref{fig:merged} shows the architecture of the Fourier Embedding Module in FM-LLM. First, the input time series tokens are projected into a high-dimensional space via a linear layer:
\[
P_k^1 = \mathrm{Linear}( P_k) \tag{9}
\]
where $P_k^1 \in \mathbb{R}^{h1}$. Then, the FAN module $\mathrm{FAN}(\cdot): \mathbb{R}^{h1}\mapsto\mathbb{R}^{h2}$ extracts frequency-related features and nonperiodic components to form the following embedding:
\[
P_k^2 = \mathrm{Tanh}(\mathrm{FAN}(\mathrm{SiLU}(P_k^1))) \tag{10}
\]
Finally, another linear layer is used to embeds the resulting representation into the hidden space of the large model:
\[
TE_k = \mathrm{Linear}(P_k^{2}), \quad k = 1, 2, \dots, N. \tag{11}
\]
This procedure constitutes the full Fourier Embedding Module of FM-LLM. This module effectively captures the dominant frequency components embedded in the original tokens, allowing the large model to utilize historical frequency information for more thorough modeling.

\subsubsection{FAN-MoE Decoder}
To enhance the expressiveness of the output module without incurring excessive computational cost, we adopt a sparse Mixture-of-Experts (MoE) architecture as the decoder in FM-LLM. MoE has been shown to effectively scale model capacity while preserving computational efficiency by activating only a small subset of expert networks for each token, based on a learned routing function \cite{du2022glam}. This allows us to maintain a lightweight inference path while expanding the model’s representation space.

Prevalent large language models~\cite{devlin2019bert,radford2019language} can predict the next token \( \mathrm{token}_k \) based on the preceding tokens \( p_{<k} \). We adopt this approach to reuse these models and iteratively generate predictions of arbitrary length. Given a time series segment of length \( NP \), the input sequence is embedded into \( N \) token embeddings \( \{ TE_1, \cdots, TE_N \} \). The training objective is to autoregressively generate the next tokens \( \{ \hat{P}_2, \cdots, \hat{P}_{N+1} \} \) based on these embeddings. We feed the token embeddings \( TE_i \) into the intermediate layers of the LLM, which model the token transitions and capture the contextual dependencies:
\[
\{\widehat{TE}_2,\cdots,\widehat{TE}_{N+1}\}=\mathrm{LLM}(\{TE_1,\cdots,TE_{N}\})\tag{12}
\]

In time-series forecasting, different input tokens may correspond to distinct temporal or frequency modes, such as trends, seasonal components, or transient anomalies. This complexity becomes even more pronounced in multivariate time series, where intricate dependencies exist not only across time but also across variables. The conditional routing property of the MoE enables structural specialization across experts, allowing the decoder to adaptively reconstruct multi-scale, nonstationary, and cross-variable patterns. Because the Fourier encoder injects frequency-aware priors into the input embeddings, the MoE decoder serves as a complementary mechanism to selectively decode these representations into the target space. This combination forms a frequency-aligned encoder-decoder pathway that improves both prediction accuracy and efficiency, especially in high-dimensional forecasting settings. These design choices are aligned with recent findings that emphasize conditional computation and specialization as effective strategies for handling complex sequence structures and variable interactions in multivariate forecasting~\cite{zhou2022fedformer}.

Here, we adopt a load-balanced MoE mechanism~\cite{liu2024deepseek} shown in Figure\ref{fig:merged}. Let $\widehat{TE}_{k+1}$ denote the output of the large language model, the predicted token is then formulated as follows:

\begin{align*}
\hat{P}_{k+1} &= \sum_{i=1}^{N_s} \text{F-FFN}_i^{(s)}(\widehat{TE}_{k+1}) + \sum_{i=1}^{N_r} g_{i,k} \text{FFN}_i^{(r)}(\widehat{TE}_{k+1}) \tag{13} \\
\text{F-FFN}_i^{(s)}(\cdot) &= \mathrm{Linear}(\mathrm{SiLU}(\mathrm{FAN}(\mathrm{Tanh}(\mathrm{Linear}(\cdot))))) \tag{14} \\
g_{i,k} &= \frac{g_{i,k}'}{\sum_{j=1}^{N_r} g_{j,k}'} \tag{15} \\
g_{i,k}' &=
\begin{cases}
s_{i,k}, & s_{i,k} \in \mathrm{Topk}\left(\{s_{j,k} \mid 1 \le j \le N_r\}, K_r\right) \\
0, & \text{otherwise}
\end{cases} \tag{16} \\
s_{i,k} &= \mathrm{Softmax}(W\widehat{TE}_{k+1}) \tag{17}
\end{align*}

Here, \( k \) represent the \( k \)-th time-series token, and \( N_s \) and \( N_r \) denote the total numbers of Fourier and routed experts, respectively. The operators \(\mathrm{F\text{-}FFN}_i^{(s)}(\cdot)\) and \(\mathrm{FFN}_i^{(r)}(\cdot)\) correspond to the \( i \)-th Fourier expert and routed expert. The variable \( g_i^{k} \) indicates the normalized gating value for the \( i \)-th routed expert on token \( k \), and \( g_{i,k} \) denotes the pre-normalized gating score; $s_{i,k}$ indicates how strongly token $k$ is associated with expert $i$; and $\mathrm{Topk}(\cdot, K_r)$ denotes the function that selects the $K_r$ highest scores among all associated values $s_{j,k}$ for the $k$-th token. $W \in \mathbb{R}^{H \times 1}$, where H denotes the LLM's hidden dimension.

It is worth noting that, to enhance the model’s ability to represent periodic patterns, we incorporate a FAN layer into the Fourier experts, similar to the encoder module, as illustrated in Fig~\ref{fig:merged}. Ensuring consistency across frequency tokens and facilitating model generalization. Meanwhile, to retain the capacity for modeling the non-periodic components of the sequence, the routed experts remain as standard feed-forward networks (FFNs) composed of two linear layers.

A prominent challenge in sparsely gated MoE models is load imbalance, where certain experts are overused while others remain underutilized, leading to routing collapse and degraded model efficiency. To mitigate the load imbalance problem in expert routing, we adopt a bias-based auxiliary-loss-free balancing mechanism inspired by DeepSeek-V3~\cite{liu2024deepseek}, where each expert is associated with a learnable bias term to adjust its selection probability during top-$K$ routing:

\begin{equation}
g'_{i,t} =
\begin{cases}
s_{i,t}, & s_{i,t} + b_i \in \mathrm{Topk}(\{s_{j,t} + b_j \mid 1 \leq j \leq N_r\}, K_r) \\
0, & \text{otherwise.}
\end{cases}
\tag{18}
\end{equation}
Through this dynamic adjustment mechanism,the FM-LLM maintained a balanced expert load during training.

\subsection{Loss Function}
Unlike previous approaches that generate a fixed-length prediction sequence in a single forward pass, autoregressive forecasting inevitably leads to error accumulation during long-horizon predictions, owing to the recursive use of model outputs as future inputs. To alleviate this problem, it is essential to enhance the single-step forecasting accuracy of the model; that, its ability to predict the next token based on the historical sequence.
To this end, a signal-decay weighting scheme inspired by CARD~\cite{xue2024card} is adopted in the loss function, where predictions closer to the present receive higher importance. This reweighting strategy encourages the model to focus on near-future predictions, thereby reducing early-stage prediction errors and indirectly mitigating the compounding effect of error accumulation in autoregressive rollouts.
Furthermore, to fully exploit the historical information contained in the input window and strengthen the model’s capability to extrapolate future trends from past observations, the training loss is computed solely based on the token immediately following the context window. The empirical results show that this design significantly improves forecasting accuracy compared to uniformly weighted multi-step objectives.
In addition, it is observed that FAN alone did not consistently enhance prediction performance when applied directly. To address this limitation, an auxiliary frequency-domain loss is introduced, which enforces an alignment between the predicted and ground-truth sequences in the Fourier domain. By minimizing the discrepancy between their Fourier coefficients, the model is encouraged to learn periodic structures more effectively, thereby allowing the FAN module to fully exploit its potential through frequency-guided supervision. The loss formula is expressed as follows:
\begin{equation}
\begin{aligned}
\mathcal{L}_{\text{forecast}} =\ & \alpha \cdot \mathcal{L}_{\text{freq}} + \beta \cdot \mathcal{L}_{\text{time}} \\
= &\ \alpha \cdot \left[
\frac{1}{L} \sum_{l=1}^{L}
\left\| \mathcal{F}( w_l\cdot\hat{a}_{t+l}) - \mathcal{F}(w_l\cdot a_{t+l}) \right\|_1
\right] \\
& + \beta \cdot \left[
\frac{1}{L} \sum_{l=1}^{L} w_l \cdot
\left\| \hat{a}_{t+l} - a_{t+l} \right\|_2^2
\right]
\end{aligned}
\tag{19}
\end{equation}

where $\mathcal{L}_{\text{forecast}}$ denotes the overall loss function, composed of two components: a frequency-domain loss and a time-domain loss. The scalar $\alpha, \beta \in [0, 1]$ is a hyperparameter that balances the contribution of each component. The summation index $l = 1, \dots, L$ iterates over $L$ future prediction steps, and in our experiments, we restrict the loss evaluation to the immediate next-token prediction, setting $L = P$ accordingly.

In the first term, $\mathcal{F}(\cdot)$ denotes the discrete Fourier transform (DFT) applied to the predicted time steps $\hat{a}_{t+l}$ and the ground truth time steps $a_{t+l}$. The $\ell_1$ norm is used to measure the discrepancy between their frequency representations. The second term computes the conventional mean squared error (MSE) in the time domain by using the $\ell_2$ norm.

The weighting coefficients $w_l = 1/\sqrt{l}$ serve as signal-decay weights, assigning larger weights to near-future predictions and smaller weights to long-term predictions. This design prioritizes accurate short-term predictions, which are crucial in autoregressive forecasting to mitigate error accumulation during long-range rollout.

Although our model adopts a primarily auxiliary-loss-free strategy for MoE expert routing, autoregressive time-series forecasting presents a unique challenge: expert load imbalance within a single prediction sequence may be amplified due to recursive dependencies. To address this issue, we introduce a complementary sequence-wise balance loss~\cite{liu2024deepseek}, defined as:

\begin{equation}
\mathcal{L}_{\text{Bal}} = \gamma \sum_{i=1}^{N_r} f_i P_i,
\tag{20}
\end{equation}
Here, \( N_r \) stands for the number of experts, and \( \gamma \) denotes a small hyperparameter responsible for adjusting the auxiliary loss weight.

The term $f_i$ denotes the normalized routing frequency of expert $i$ across all time series tokens, which is computed as:
\begin{equation}
f_i = \frac{N_r}{K_r K} \sum_{k=1}^{K} \mathds{1} (s_{i,k} \in \text{Top}k(\{s_{j,k}\}_{j=1}^{N_r}, K_r)),
\tag{21}
\end{equation}
where $K_r$ is the number of experts selected per token, $s_{i,k}$ is the routing score of expert $i$ for the $k$-th token, and $\mathds{1}(\cdot)$ is the indicator function.

We further define the normalized routing proportion $s_{i,k}'$ and the averaged routing probability $P_i$ as:
\begin{equation}
s_{i,k}' = \frac{s_{i,k}}{\sum_{j=1}^{N_r} s_{j,k}},
\tag{22}
\end{equation}

\begin{equation}
P_i = \frac{1}{T} \sum_{t=1}^{T} s_{i,k}'.
\tag{23}
\end{equation}

This sequence-wise auxiliary loss encourages token-wise expert balance within each sequence, thereby preventing local routing collapse in autoregressive prediction settings. In practice, the coefficient $\lambda$ is set to a small value, ensuring that it does not dominate the main forecasting objectives while still offering regularization benefits.

Accordingly, the overall loss function is defined as follows:
\begin{equation}
\mathcal{L}_{\text{total}} =
\underbrace{
\alpha \cdot \mathcal{L}_{\text{freq}} +
\beta \cdot \mathcal{L}_{\text{time}}
}_{\text{Forecasting Loss}}
+ 
\lambda \cdot \mathcal{L}_{\text{Bal}}
\tag{24}
\end{equation}
This multi-objective loss design ensures that the model not only achieves accurate and robust time-series prediction but also maintains stable and efficient expert routing behavior during long-horizon recursive forecasting.

\section{Experiments}
\subsection{Experimental setup}
\subsubsection{Datasets}

\begin{table}[!b]
    \centering
    \caption{Dataset detailed descriptions. The dataset size is organized in (Train, Validation, Test).The forecastability is calculated by one minus the entropy of Fourier decomposition of time series~\cite{pmlr-v28-goerg13}.A larger value indicates better predictability.}
    \label{tab:data_stats}
    \resizebox{\textwidth}{!}{
        \begin{tabular}{c|l|c|c|c|c|c|c}
            \toprule 
            Tasks & Dataset & Dim & Series Length & Dataset Size & Frequency & Forecastability* & Information \\
            \midrule 
            \multirow{11}{*}{\shortstack{Long-term\\Forecasting}} 
              & ETTm1 & 7 & \{96, 192, 336, 720\} & (34465, 11521, 11521) & 15min & 0.46 & Temperature \\
            \cmidrule{2-8} 
              & ETTm2 & 7 & \{96, 192, 336, 720\} & (34465, 11521, 11521) & 15min & 0.55 & Temperature \\
            \cmidrule{2-8}
              & ETTh1 & 7 & \{96, 192, 336, 720\} & (8545, 2881, 2881) & 15 min & 0.38 & Temperature \\
            \cmidrule{2-8}
              & ETTh2 & 7 & \{96, 192, 336, 720\} & (8545, 2881, 2881) & 15 min & 0.45 & Temperature \\
            \cmidrule{2-8}
              & Electricity & 321 & \{96, 192, 336, 720\} & (18317, 2633, 5261) & Hourly & 0.77 & Electricity \\
            \cmidrule{2-8}
              & Traffic & 862 & \{96, 192, 336, 720\} & (12185, 1757, 3509) & Hourly & 0.68 & Transportation \\
            \cmidrule{2-8}
              & Weather & 21 & \{96, 192, 336, 720\} & (36792, 5271, 10540) & 10 min & 0.75 & Weather \\
            \cmidrule{2-8}
              & PEMS03 & 358 & 12 & (15617, 5135, 5135) & 5min & 0.65 & Transportation \\
            \cmidrule{2-8}
              & PEMS04 & 307 & 12 & (10172, 3375, 3375) & 5min & 0.45 & Transportation \\
            \cmidrule{2-8}
              & PEMS07 & 883 & 12 & (16911, 5622, 5622) & 5min & 0.58 & Transportation \\
            \cmidrule{2-8}
              & PEMS08 & 170 & 12 & (10690, 3548, 265) & 5min & 0.52 & Transportation \\
            \midrule 
            \multirow{6}{*}{\shortstack{Short-term\\Forecasting}} 
              & M4-Yearly & 1 & 6 & (23000, 0, 23000) & Yearly & 0.43 & Demographic \\
            \cmidrule{2-8}
              & M4-Quarterly & 1 & 8 & (24000, 0, 24000) & Quarterly & 0.47 & Finance \\
            \cmidrule{2-8}
              & M4-Monthly & 1 & 18 & (48000, 0, 48000) & Monthly & 0.44 & Industry \\
            \cmidrule{2-8}
              & M4-Weakly & 1 & 13 & (359, 0, 359) & Weakly & 0.43 & Macro \\
            \cmidrule{2-8}
              & M4-Daily & 1 & 14 & (4227, 0, 4227) & Daily & 0.44 & Micro \\
            \cmidrule{2-8}
              & M4-Hourly & 1 & 48 & (414, 0, 414) & Hourly & 0.46 & Other \\
            \bottomrule 
        \end{tabular}
    }
\end{table}

We assess the long-term forecasting capabilities of FM-LLM on seven distinct benchmark datasets, including the ETT datasets (namely ETTh1, ETTh2, ETTm1, and ETTm2), alongside Electricity, Weather, and Traffic datasets.
The Electricity Transformer Temperature (ETT) benchmark dataset acts as a key indicator for modeling long-term electric power deployment trends. It contains two years of operational data gathered from two counties in China and is divided into four subsets featuring different temporal resolutions: ETTh1 and ETTh2, sampled hourly, and ETTm1 and ETTm2, sampled every 15 minutes. Each record consists of six power load-related features and a target variable known as oil temperature, which represents the transformer's thermal condition and is widely utilized in predictive modeling.
The Electricity dataset comprises hourly electricity consumption data from 321 individual consumers, providing a high-dimensional multivariate time series capturing daily and weekly usage patterns.
The Weather dataset includes one year of meteorological data collected from 21 monitoring stations throughout Germany, with a sampling frequency of every 10 minutes. It covers a broad range of atmospheric variables that facilitate fine-grained forecasting and environmental modeling.
The Traffic dataset contains hourly occupancy rates from 862 sensors distributed across the freeway network in California. This dataset captures temporal traffic patterns and is frequently used in spatiotemporal modeling and congestion prediction research.
The PEMS dataset consists of public traffic network data from California, collected in 5-minute intervals. We use the same four public subsets (PEMS03, PEMS04, PEMS07,PEMS08) adopted in iTransformer~\cite{liu2023itransformer}.The details are provided in Table~\ref{tab:data_stats}.

Following standard practice, we partition all datasets in chronological order. Specifically, the ETT datasets are split into training, validation, and test sets with a ratio of (3:1:1), the PEMS datasets adopt a (6:2:2) split, whereas the other datasets adopt a (7:1:2) split.

\subsubsection{Metrics}
For all experiments, we adopt the same evaluation criteria as used in mainstream methods. Specifically, for long-term forecasting tasks, we employ Mean Squared Error and Mean Absolute Error as performance evaluation metrics. For the short-term forecasting, following the~\cite{wu2023timesnet}, we adopt the symmetric mean absolute percentage error (SMAPE), mean absolute scaled error (MASE) and overall weighted average (OWA) as the metrics, where OWA is a special metric used in M4 competition. The definitions of these metrics are as follows:
\begin{align}
\text{MSE} &= \frac{1}{N} \sum_{i=1}^{N} (y_i - \hat{y}_i)^2 \tag{25}\\
\text{MAE} &= \frac{1}{N} \sum_{i=1}^{N} |y_i - \hat{y}_i|\tag{26} \\
\text{SMAPE} &= \frac{200}{N} \sum_{i=1}^{N} \frac{|y_i - \hat{y}_i|}{|y_i| + |\hat{y}_i|} \tag{27} \\
\text{MASE} &= \frac{1}{N} \sum_{i=1}^{N} \frac{|y_i - \hat{y}_i|}{\frac{1}{N-s} \sum_{j=s+1}^{N} |y_j - y_{j-s}|} \tag{28} \\
\text{OWA} &= \frac{1}{2} \left[ \frac{\text{SMAPE}}{\text{SMAPE}_{\text{Na\"ive2}}} + \frac{\text{MASE}}{\text{MASE}_{\text{Na\"ive2}}} \right] \tag{29} 
\end{align}
where $N$ denotes the number of data points(i.e, the length of one token, which is equal to $P$), $y_i$ and $\hat{y}_i$ are the $i$-th groundtruth and prediction where $i \in \{1,2,\cdots,N\}$, s is periodicity of the data.

\subsubsection{Baselines}
We carefully select 16 representative baselines from the recent time series forecasting landscape, including the following categories: (1) Transformer-based models:  iTransformer~\cite{liu2023itransformer}, PatchTST~\cite{Yuqietal-2023-PatchTST}, TQ-Net~\cite{lin2025TQNet}, Non-Stationary Transformer~\cite{liu2022nonstationary}; Reformer~\cite{kitaev2020reformer}(2) MLP-based models: DLinear~\cite{Zeng_Chen_Zhang_Xu_2023}, AMD~\cite{hu2025adaptive}, CycleNet~\cite{lin2024cyclenet}, N-Beats~\cite{oreshkin2020nbeats}, N-Hits~\cite{challu2023nhits};
(3) CNN-based models: Timesnet~\cite{wu2023timesnet}, FEDformer~\cite{zhou2022fedformer}; 
(4) LLM-based models: AutoTimes~\cite{liu2024autotimes}, GPT4TS~\cite{zhou2023one}, PRADA~\cite{LIU2025113478}, TimeLLM~\cite{jin2023time}.

\begin{table*}[!b]
\centering
\small 
\caption{Model configurations of FM-LLM. Fourier refer to Fourier experts}
\begin{adjustbox}{max width=\textwidth}
\begin{tabular}{ccccccccccccc}
\toprule
Dataset & token len & MLP dim & dropout & lr & Fourier & routed & active & $\gamma$ & $\alpha$ & $\beta$ & $\lambda$ & batch \\
\midrule
ETTm1 & 96 & 512 & 0.2 & 2e-4 & 2 & 6 & 2 & 1e-4 & 1 & 1 & 1 & 256 \\
ETTm2 & 96 & 512 & 0.2 & 1.5e-4 & 2 & 6 & 2 & 1e-4 & 1 & 1 & 1 & 256 \\
ETTh1 & 96 & 512 & 0.2 & 2e-4 & 2 & 2 & 1 & 1e-4 & 1 & 1 & 1 & 256 \\
ETTh2 & 96 & 512 & 0.2 & 2e-4 & 2 & 3 & 1 & 1e-4 & 1 & 1 & 1 &256 \\
Weather & 96 & 512 & 0.1 & 1e-4 & 2 & 8 & 3 & 1e-4 & 1 & 1 & 1 & 256 \\
Electricity & 96 & 512 & 0.1 & 1e-4 & 2 & 8 & 3 & 1e-5 & 0.9 & 1 & 0.1 &256 \\
Traffic & 96 & 512 & 0.1 & 1e-4 & 2 & 8 & 3 & 1e-5 & 0.9 & 1 & 0.1 &256 \\
PEMS03 & 96 & 512 & 0.1 & 1e-4 & 2 & 8 & 3 & 1e-5 & 0.9 & 1 & 0.1 &256 \\
PEMS04 & 96 & 512 & 0.1 & 1e-4 & 2 & 8 & 3 & 1e-5 & 0.9 & 1 & 0.1 &256 \\
PEMS07 & 96 & 512 & 0.1 & 1e-4 & 2 & 8 & 3 & 1e-5 & 0.9 & 1 & 0.1 &256 \\
PEMS08 & 96 & 512 & 0.1 & 1e-4 & 2 & 8 & 3 & 1e-5 & 0.9 & 1 & 0.1 &256 \\
M4 Yearly & 6 & 256 & 0.1 & 1e-4 & 2 & 4 & 2 & 1e-4 & 1 & 1 & 1 & 64 \\
M4 Quarterly & 8 & 512 & 0.2 & 5e-5 & 2 & 6 & 2 & 1e-4 & 1 & 1 & 1 & 16 \\
M4 Monthly & 18 & 512 & 0.0 & 5e-5 & 2 & 2 & 1 & 1e-4 & 1 & 1 & 1 & 16 \\
M4 Weekly & 13 & 256 & 0.0 & 1e-3 & 2 & 2 & 1 & 1e-4 & 1 & 1 & 1 & 16 \\
M4 Daily & 14 & 512 & 0.0 & 5e-4 & 2 & 2 & 1 & 1e-4 & 1 & 1 & 1 & 16 \\
M4 Hourly & 48 & 256 & 0.0 & 2e-4 & 2 & 2 & 1 & 1e-4 & 1 & 1 & 1 & 16 \\
\bottomrule
\end{tabular}
\end{adjustbox}

\label{tab:FM-LLM_config}
\end{table*}

\subsubsection{Implementation Details}
For a fair comparison, all the experiments share the same random
 seed and use the Adam optimizer. We use Llama-3.2-1B as our backbone. We employ the L1 loss as the optimization objective for the ETT, traffic and Weather datasets, whereas the MSE loss is used for Electricity datasets. The hyperparameters were tuned manually based on validation set performance, the early stopping criterion used a patience of 3. Our model implementation is based on PyTorch 2.5.1~\cite{paszke2019pytorch}, with all experiments conducted on a single NVIDIA RTX 4090 24G GPU. The training process demands approximately 6GB of GPU memory and involves roughly 8.68 million trainable parameters.

In addition, we follow the same settings of lookback window length and token length as in AutoTimes~\cite{liu2024autotimes}. Specifically, for our model, we set the input length to $L=672$ and the output token length to $P=96$. To ensure fair and consistent comparison, we primarily use the results reported in the original publications for all baseline models. For the PEMS datasets, we set the input length of each method to 672. If the forecasting performance at this length does not exceed the results reported in the original paper, we retain the results from the original publication. In the few-shot learning setting, we train the model for up to 100 epochs with an early stopping patience of 10, to ensure effective fitting on limited data. For short-term forecasting tasks, we set the lookback window size to twice the prediction horizon. Other details of forecasting configurations are summarized in Table~\ref{tab:FM-LLM_config}.

\subsection{Main Results}
Table~\ref{tab:benchmark-results-avg}, 
Table~\ref{tab:m4-average}, 
Table~\ref{tab:PEMS}, 
Table~\ref{tab:example_results}, and Table~\ref{tab:transfer_results} summarize the main results of FM-LLM on multivariate time-series forecasting tasks, including Long-term forecasting, Short-term forecasting, few-shot learning, and zero-shot transfer experiments. Overall, the results indicate that FM-LLM not only achieves significantly superior performance in standard long-sequence forecasting but also demonstrates outstanding adaptability and stability under generalization challenges such as data scarcity (few-shot) and cross-distribution transfer (zero-shot).

\begin{table*}[!b]
\centering

\begin{minipage}{\textwidth}
\centering
\captionsetup{type=table}
\caption{Long-term forecasting results. We average the results across 4 prediction lengths: \{96, 192, 336, 720\}. The best results are highlighted in \best{bold red} and the second best are \second{underlined blue}. \hyperref[app1]{Appendix~A} provides full results.}
\label{tab:benchmark-results-avg}

{\scriptsize
\renewcommand{\arraystretch}{1.35}
\setlength{\tabcolsep}{2.8pt}
\setlength{\extrarowheight}{0.6pt}

\resizebox{\textwidth}{!}{
\begin{tabular}{l*{22}{c}}  
\toprule

\multicolumn{1}{c|}{\textbf{Models}}
& \multicolumn{2}{c}{\makecell{\textbf{FM-LLM}\\\textbf{(Ours)}}}
& \multicolumn{2}{c}{\makecell{\textbf{AutoTimes*}\\\cite{liu2024autotimes}}}
& \multicolumn{2}{c}{\makecell{\textbf{GPT4TS}\\\cite{zhou2023one}}}
& \multicolumn{2}{c}{\makecell{\textbf{PRADA}\\\cite{LIU2025113478}}}
& \multicolumn{2}{c}{\makecell{\textbf{PatchTST}\\\cite{Yuqietal-2023-PatchTST}}}
& \multicolumn{2}{c}{\makecell{\textbf{iTransformer}\\\cite{liu2023itransformer}}}
& \multicolumn{2}{c}{\makecell{\textbf{MICN}\\\cite{wang2023micn}}}
& \multicolumn{2}{c}{\makecell{\textbf{TimesNet}\\\cite{wu2023timesnet}}}
& \multicolumn{2}{c}{\makecell{\textbf{FEDformer}\\\cite{zhou2022fedformer}}}
& \multicolumn{2}{c}{\makecell{\textbf{DLinear}\\\cite{Zeng_Chen_Zhang_Xu_2023}}}
& \multicolumn{2}{c}{\makecell{\textbf{AMD}\\\cite{hu2025adaptive}}}
\\

\cmidrule(lr){2-3}\cmidrule(lr){4-5}\cmidrule(lr){6-7}\cmidrule(lr){8-9}\cmidrule(lr){10-11}
\cmidrule(lr){12-13}\cmidrule(lr){14-15}\cmidrule(lr){16-17}\cmidrule(lr){18-19}\cmidrule(lr){20-21}\cmidrule(lr){22-23}

\multicolumn{1}{c|}{\textbf{Metric}}
& \textbf{MSE} & \multicolumn{1}{c|}{\textbf{MAE}}
& \textbf{MSE} & \multicolumn{1}{c|}{\textbf{MAE}}
& \textbf{MSE} & \multicolumn{1}{c|}{\textbf{MAE}}
& \textbf{MSE} & \multicolumn{1}{c|}{\textbf{MAE}}
& \textbf{MSE} & \multicolumn{1}{c|}{\textbf{MAE}}
& \textbf{MSE} & \multicolumn{1}{c|}{\textbf{MAE}}
& \textbf{MSE} & \multicolumn{1}{c|}{\textbf{MAE}}
& \textbf{MSE} & \multicolumn{1}{c|}{\textbf{MAE}}
& \textbf{MSE} & \multicolumn{1}{c|}{\textbf{MAE}}
& \textbf{MSE} & \multicolumn{1}{c|}{\textbf{MAE}}
& \textbf{MSE} & \textbf{MAE}
\\
\midrule

\multicolumn{1}{c|}{ETTm1}
& \bestx{0.338} & \multicolumn{1}{c|}{\bestx{0.369}}
& 0.356 & \multicolumn{1}{c|}{0.386}
& 0.352 & \multicolumn{1}{c|}{0.383}
& \secondx{0.341} & \multicolumn{1}{c|}{0.376}
& 0.351 & \multicolumn{1}{c|}{0.381}
& 0.373 & \multicolumn{1}{c|}{0.403}
& 0.387 & \multicolumn{1}{c|}{0.411}
& 0.400 & \multicolumn{1}{c|}{0.406}
& 0.448 & \multicolumn{1}{c|}{0.452}
& 0.362 & \multicolumn{1}{c|}{0.379}
& 0.347 & \second{0.374}
\\

\multicolumn{1}{c|}{ETTm2}
& \secondx{0.253} & \multicolumn{1}{c|}{\bestx{0.303}}
& 0.275 & \multicolumn{1}{c|}{0.326}
& 0.266 & \multicolumn{1}{c|}{0.326}
& \bestx{0.250} & \multicolumn{1}{c|}{\secondx{0.309}}
& 0.255 & \multicolumn{1}{c|}{0.315}
& 0.274 & \multicolumn{1}{c|}{0.336}
& 0.284 & \multicolumn{1}{c|}{0.340}
& 0.291 & \multicolumn{1}{c|}{0.333}
& 0.305 & \multicolumn{1}{c|}{0.349}
& 0.256 & \multicolumn{1}{c|}{0.331}
& 0.254 & 0.315
\\

\multicolumn{1}{c|}{ETTh1}
& \bestx{0.378} & \multicolumn{1}{c|}{\bestx{0.407}}
& \secondx{0.393} & \multicolumn{1}{c|}{0.424}
& 0.427 & \multicolumn{1}{c|}{0.426}
& 0.402 & \multicolumn{1}{c|}{\secondx{0.422}}
& 0.413 & \multicolumn{1}{c|}{0.431}
& 0.438 & \multicolumn{1}{c|}{0.450}
& 0.440 & \multicolumn{1}{c|}{0.462}
& 0.458 & \multicolumn{1}{c|}{0.450}
& 0.440 & \multicolumn{1}{c|}{0.460}
& 0.423 & \multicolumn{1}{c|}{0.437}
& 0.407 & 0.424
\\

\multicolumn{1}{c|}{ETTh2}
& \secondx{0.336} & \multicolumn{1}{c|}{\bestx{0.378}}
& 0.356 & \multicolumn{1}{c|}{0.399}
& 0.354 & \multicolumn{1}{c|}{0.394}
& 0.348 & \multicolumn{1}{c|}{0.391}
& \bestx{0.330} & \multicolumn{1}{c|}{0.379}
& 0.367 & \multicolumn{1}{c|}{0.407}
& 0.402 & \multicolumn{1}{c|}{0.437}
& 0.414 & \multicolumn{1}{c|}{0.427}
& 0.437 & \multicolumn{1}{c|}{0.449}
& 0.431 & \multicolumn{1}{c|}{0.447}
& 0.351 & 0.392
\\

\multicolumn{1}{c|}{Traffic}
& \bestx{0.377} & \multicolumn{1}{c|}{\bestx{0.249}}
& 0.394 & \multicolumn{1}{c|}{0.272}
& 0.414 & \multicolumn{1}{c|}{0.294}
& 0.395 & \multicolumn{1}{c|}{0.278}
& 0.391 & \multicolumn{1}{c|}{\secondx{0.264}}
& \secondx{0.380} & \multicolumn{1}{c|}{0.272}
& 0.542 & \multicolumn{1}{c|}{0.316}
& 0.620 & \multicolumn{1}{c|}{0.336}
& 0.610 & \multicolumn{1}{c|}{0.376}
& 0.434 & \multicolumn{1}{c|}{0.295}
& 0.468 & 0.271
\\

\multicolumn{1}{c|}{Electricity}
& \bestx{0.156} & \multicolumn{1}{c|}{\bestx{0.246}}
& 0.164 & \multicolumn{1}{c|}{0.257}
& 0.167 & \multicolumn{1}{c|}{0.263}
& 0.162 & \multicolumn{1}{c|}{0.258}
& 0.159 & \multicolumn{1}{c|}{0.253}
& 0.177 & \multicolumn{1}{c|}{0.274}
& 0.187 & \multicolumn{1}{c|}{0.295}
& 0.192 & \multicolumn{1}{c|}{0.295}
& 0.214 & \multicolumn{1}{c|}{0.327}
& 0.177 & \multicolumn{1}{c|}{0.224}
& \secondx{0.157} & \secondx{0.250}
\\

\multicolumn{1}{c|}{Weather}
& \secondx{0.225} & \multicolumn{1}{c|}{\bestx{0.255}}
& 0.238 & \multicolumn{1}{c|}{0.271}
& 0.237 & \multicolumn{1}{c|}{0.270}
& \secondx{0.223} & \multicolumn{1}{c|}{\secondx{0.261}}
& 0.226 & \multicolumn{1}{c|}{0.264}
& 0.238 & \multicolumn{1}{c|}{0.273}
& 0.243 & \multicolumn{1}{c|}{0.299}
& 0.259 & \multicolumn{1}{c|}{0.287}
& 0.309 & \multicolumn{1}{c|}{0.360}
& 0.240 & \multicolumn{1}{c|}{0.300}
& \bestx{0.222} & 0.261
\\

\midrule
\multicolumn{1}{c|}{\textbf{1st Count}}
& \multicolumn{2}{c|}{\textbf{11}}
& \multicolumn{2}{c|}{0}
& \multicolumn{2}{c|}{0}
& \multicolumn{2}{c|}{\textbf{1}}
& \multicolumn{2}{c|}{\textbf{1}}
& \multicolumn{2}{c|}{0}
& \multicolumn{2}{c|}{0}
& \multicolumn{2}{c|}{0}
& \multicolumn{2}{c|}{0}
& \multicolumn{2}{c|}{0}
& \multicolumn{2}{c}{\textbf{1}}
\\

\bottomrule
\end{tabular}
}
} 
\end{minipage}

\vspace{6pt}

\begin{minipage}{\textwidth}
\centering
\captionsetup{type=table}
\caption{Short-term time series forecasting results on M4. The forecasting horizons are in [6, 48] and the three rows provided are weighted averaged from all datasets under different sampling intervals. A lower value indicates better performance. Red: the best, Blue: the second best. \hyperref[app1]{Appendix~A} provides full results.}
\label{tab:m4-average}

{\scriptsize
\renewcommand{\arraystretch}{1.35}
\setlength{\extrarowheight}{1.1pt}
\setlength{\tabcolsep}{4.8pt}

\resizebox{\textwidth}{!}{
\begin{tabular}{cc*{11}{c}}
\toprule

\multicolumn{2}{c|}{\textbf{Methods}}
& \makecell{\textbf{FM-LLM}}
& \makecell{\textbf{TIMELLM}}
& \makecell{\textbf{GPT4TS}}
& \makecell{\textbf{TimesNet}}
& \makecell{\textbf{PatchTST}}
& \makecell{\textbf{N-HiTS}}
& \makecell{\textbf{N-BEATS}}
& \makecell{\textbf{DLinear}}
& \makecell{\textbf{FEDformer}}
& \makecell{\textbf{Stationary}}
& \makecell{\textbf{Reformer}}
\\

\cmidrule(lr){1-2}\cmidrule(lr){3-13}

\multicolumn{1}{c}{\multirow{3}{*}{\rotatebox[origin=c]{90}{\textbf{Average}}}}
& \multicolumn{1}{c|}{\textbf{SMAPE}}
& \bestx{11.840}
& \secondx{11.983}
& 12.69
& 12.88
& 12.059
& 12.035
& 12.25
& 13.639
& 13.16
& 12.780
& 18.200
\\

& \multicolumn{1}{c|}{\textbf{MASE}}
& \bestx{1.585}
& \secondx{1.595}
& 1.808
& 1.836
& 1.623
& 1.625
& 1.698
& 2.095
& 1.775
& 1.756
& 4.223
\\

& \multicolumn{1}{c|}{\textbf{OWA}}
& \bestx{0.851}
& \secondx{0.859}
& 0.94
& 0.955
& 0.869
& 0.869
& 0.896
& 1.051
& 0.949
& 0.930
& 1.775
\\

\bottomrule
\end{tabular}
}
}
\end{minipage}

\end{table*}

Table~\ref{tab:benchmark-results-avg} summarizes the long-term forecasting results by averaging over four prediction lengths. For completeness, we report the full results in Appendix~A Table~\ref{tab:m4-full-breakdown}, where the best results are marked in bold and the second-best are underlined. Across the 70 reported metrics, FM-LLM achieves 51 best and 15 second-best scores, demonstrating a strong overall advantage. Notably, AutoTimes, which also adopts an autoregressive architecture and exhibits good transfer generalization, is consistently outperformed by FM-LLM across all seven datasets for long-sequence forecasting. On average, FM-LLM reduces MSE and MAE by 5.20\% and 5.32\%, respectively, compared to AutoTimes. This clearly indicates that the Frequency-Aware Module (Fourier Analysis Network, FAN) and the time-frequency joint loss function introduced in FM-LLM effectively enhance the global modeling capacity of the autoregressive forecasting framework and improve its ability to capture long-range dependencies.

From the perspective of multivariate forecasting, FM-LLM also maintains its advantage on high-dimensional datasets such as Electricity (321 variables) and Traffic (862 variables), significantly outperforming most baseline models. We attribute this advantage to the FAN-MoE decoder architecture, which substantially enhances the expressive power compared to conventional single-layer MLPs. In multivariate settings, the type and combinatorial complexity of input tokens increase dramatically, making it difficult for simple linear projections or MLP decoders to effectively allocate and model such highly heterogeneous information. FM-LLM’s MoE decoder leverages a dynamic routing mechanism to activate the most suitable expert for each subsequence structure, thereby improving the model’s ability to express weak inter-variable correlations and asynchronous periodic patterns, resulting in superior performance.

We further evaluate FM-LLM on the M4 benchmark for short-term forecasting, where the forecasting horizons fall in $[6,48]$.
Following common practice, we report SMAPE/MASE/OWA and provide weighted averages over all series under different sampling intervals.
As shown in Table~\ref{tab:m4-average}, FM-LLM achieves the best overall performance, and the full breakdown is deferred to \hyperref[app1]{Appendix~A} provides full results.

Table~\ref{tab:PEMS} presents the forecasting performance on the PEMS datasets (PEMS03, PEMS04, PEMS07, PEMS08), with results averaged over four prediction lengths. Compared to TQ-Net, CycleNet, and iTransformer, the proposed FM-LLM (ours) achieves the lowest MSE and MAE across all datasets. Specifically, FM-LLM outperforms the second-best method by an average of 6\% in MSE and 7.5\% in MAE, further validating its effectiveness on large-scale multivariate traffic datasets.

\begin{table*}[!b]
\begin{minipage}{\linewidth}
    
\end{minipage}
  \centering
  \caption{Forecasting performance on the PEMS datasets.The results are averaged from all four prediction lengths.The best results are in \best{bold} and the second best are \second{underlined}.}
  \label{tab:PEMS}
  \begin{adjustbox}{max width=\textwidth}

  \begin{tabular}{c|cc|cc|cc|cc}
  
    \toprule
    \multirow{2}{*}{\textbf{Models}}
      & \multicolumn{2}{c|}{PEMS03}
      & \multicolumn{2}{c|}{PEMS04}
      & \multicolumn{2}{c|}{PEMS07}
      & \multicolumn{2}{c}{PEMS08} \\
      & \textbf{MSE} & \textbf{MAE}
      & \textbf{MSE} & \textbf{MAE}
      & \textbf{MSE} & \textbf{MAE}
      & \textbf{MSE} & \textbf{MAE} \\
    \midrule


    TQNet~\cite{lin2025TQNet}   & \second{0.097} & 0.203
               & 0.091 & 0.197
               & 0.075 & 0.171
               & 0.142 & 0.229 \\

    CycleNet~\cite{lin2024cyclenet}   & 0.098 & \second{0.201}
               & \second{0.090} & \second{0.196}
               & \second{0.071} & \second{0.170}
               & \second{0.120} & \second{0.198} \\
               
    iTransformer~\cite{liu2023itransformer} & 0.098 & 0.203
                & 0.097 & 0.199
                & 0.139 & 0.245
                & 0.150 & 0.226 \\

    FM-LLM(ours)    & \best{0.089} & \best{0.186}
               & \best{0.089} & \best{0.187}
               & \best{0.063} & \best{0.151}
               & \best{0.118} & \best{0.184} \\

    \bottomrule
  \end{tabular}
  \end{adjustbox}
\end{table*}

\begin{table*}[!b]
\begin{minipage}{\linewidth}
  \centering
  \caption{Few-shot learning results on 10\% training data of
ETT datasets. The best results are in \best{bold} and the second best are \second{underlined}.}
  \label{tab:example_results}
\begin{adjustbox}{max width=\textwidth}
  \begin{tabular}{c|cc|cc|cc|cc}
    \toprule
    \multirow{2}{*}{\textbf{Models}}
      & \multicolumn{2}{c|}{\textbf{ETTm1}}
      & \multicolumn{2}{c|}{\textbf{ETTm2}}
      & \multicolumn{2}{c|}{\textbf{ETTh1}}
      & \multicolumn{2}{c}{\textbf{ETTh2}} \\
      & \textbf{MSE} & \textbf{MAE}
      & \textbf{MSE} & \textbf{MAE}
      & \textbf{MSE} & \textbf{MAE}
      & \textbf{MSE} & \textbf{MAE} \\
    \midrule

    DLinear      & \best{0.411} & \best{0.429}   & 0.316 & 0.368   & 0.647 & 0.552   & 0.605 & 0.538 \\
    TimesNet     & 0.673 & 0.534   & 0.321 & 0.354   & 0.869 & 0.628   & 0.479 & 0.465 \\
    FEDformer    & 0.696 & 0.572   & 0.356 & 0.392   & 0.639 & 0.561   & 0.466 & 0.475 \\
    iTransformer & 0.728 & 0.565   & 0.336 & 0.373   & 0.910 & 0.860   & 0.489 & 0.415 \\
    PatchTST     & 0.501 & 0.466   & 0.296 & 0.343   & 0.633 & 0.542   & 0.415 & 0.431 \\
    GPT4TS       & 0.461 & 0.441   & 0.293 & 0.335   & 0.590 & 0.525   & 0.397 & 0.421 \\
    AutoTimes    & 0.611 & 0.508   & 0.289 & 0.339   & 0.592 & 0.549   & \second{0.372} & \second{0.418}\\
    PRADA        & \second{0.435} & \second{0.428}   & \second{0.281} & \second{0.329}   & \second{0.557} & \second{0.515}   & 0.384 & 0.419 \\
    FM-LLM      & 0.457 & 0.445
                 & \best{0.259} & \best{0.316}
                 & \best{0.489} & \best{0.482}
                 & \best{0.348} & \best{0.399} \\
    \bottomrule
  \end{tabular}
  \end{adjustbox}
  \end{minipage}
\end{table*}


\begin{table*}[!b]
\begin{minipage}{\linewidth}
    
\end{minipage}
  \centering
  \caption{Zero-shot learning results on ETT datasets. A $\to$ B means that we train on dataset A while testing the model performance on dataset B. The best results are in \best{bold} and the second best are \second{underlined}.}
  \label{tab:transfer_results}
  \begin{adjustbox}{max width=\textwidth}

  \begin{tabular}{c|cc|cc|cc|cc}
  
    \toprule
    \multirow{2}{*}{\textbf{Models}}
      & \multicolumn{2}{c|}{h1 $\to$  \text{h2}}
      & \multicolumn{2}{c|}{h1 $\to$ \text{m2}}
      & \multicolumn{2}{c|}{h2 $\to$  \text{h1}}
      & \multicolumn{2}{c}{h2 $\to$ \text{m2}} \\
      & \textbf{MSE} & \textbf{MAE}
      & \textbf{MSE} & \textbf{MAE}
      & \textbf{MSE} & \textbf{MAE}
      & \textbf{MSE} & \textbf{MAE} \\
    \midrule


    DLinear    & 0.493 & 0.488
               & 0.415 & 0.452
               & 0.703 & 0.574
               & 0.328 & 0.386 \\

    TimesNet   & 0.421 & 0.431
               & 0.327 & 0.361
               & 0.865 & 0.621
               & 0.342 & 0.376 \\
               
    PatchTST   & 0.380 & 0.405
               & 0.597 & 0.483
               & 0.565 & 0.513
               & 0.325 & 0.365 \\

    GPT4TS     & 0.406 & 0.422
               & 0.325 & 0.363
               & 0.757 & 0.578
               & 0.335 & 0.370 \\
    AutoTimes  & 0.355 & 0.394
               & 0.347 & 0.374
               & 0.565 & 0.529
               & 0.322 & 0.369\\

    PRADA      & \second{0.350} & \second{0.392}
               & \second{0.307} & \second{0.357}
               & \best{0.444} & \second{0.457}
               & \second{0.296} & \second{0.349} \\

    FM-LLM    & \best{0.349} & \best{0.384}
               & \best{0.297} & \best{0.349}
               & \second{0.448} & \best{0.455}
               & \best{0.283} & \best{0.340} \\

    \bottomrule
  \end{tabular}
  \end{adjustbox}
\end{table*}

As shown in Table~\ref{tab:example_results} under the few-shot setting, FM-LLM achieves performance that even surpasses some fully supervised baselines using only the top 10\% of the training data, with particularly strong results on the ETTh1 and ETTh2 datasets. In contrast, several models with larger parameter sizes (e.g., FEDformer, TimesNet, or AutoTimes) typically suffer performance degradation under limited sample conditions and fail to extract stable and effective temporal features. FM-LLM’s advantage mainly stems from the strong structural inductive bias introduced by its frequency-enhanced encoder and expert-routing decoder, enabling robust learning of periodic structures and key dynamic features even in low-sample regimes and showcasing excellent few-shot generalization.

Table~\ref{tab:transfer_results} further investigates the zero-shot transfer performance of FM-LLM under the setting where training and testing are conducted on entirely different datasets. In this setup, the model is trained solely on a source dataset and directly applied to a target dataset at test time without any fine-tuning. FM-LLM achieves the best results across all four transfer scenarios (e.g., ETTh1 → ETTh2, ETTm1 → ETTm2), demonstrating high robustness under task transfer, distribution shift, and variable structural changes. This capability is partly attributed to the frequency-space representations modeled by FM-LLM, which, via the FAN module, capture the intrinsic periodicity and dynamic patterns of input sequences and naturally support cross-domain transfer. Simultaneously, the MoE decoder flexibly activates the appropriate feature pathways when handling data from different sources, thereby adapting to input structural variations and ensuring stable outputs.

In comparison, some representative large language model paradigms for time series (LLM4TS), such as PRADA and AutoTimes, also exhibit certain advantages in zero-shot and few-shot settings but follow different methodological paths than FM-LLM. AutoTimes adopts a tokenization mechanism to convert time series into language-like sequences and uses a frozen language model backbone with trainable projection layers for transfer learning. While this approach benefits from the strong contextual modeling capacity of LLMs and the natural generalization of in-context learning, it faces limitations: (1) token representations are task-sensitive and prone to distortion under significant frequency structure changes; (2) the large model size hinders deployment in industrial or edge scenarios. PRADA, by introducing modality alignment and dynamic adaptation mechanisms, preserves pretraining advantages while enabling task-specific adaptation and achieves solid zero-shot performance, but its explicit modeling of input distribution shifts is still less intuitive and stable compared to FM-LLM’s frequency decomposition approach.

In summary, FM-LLM incorporates a Fourier Analysis Network (FAN) in the encoder to model subsequence frequency structures, adopts an MoE module in the decoder to enhance expressive diversity, and introduces a time-frequency hybrid loss function to improve long-range modeling precision and cross-task generalization. It achieves state-of-the-art performance across supervised, few-shot, and zero-shot settings. Compared to existing large-model paradigms, FM-LLM demonstrates a favorable balance between interpretability, adaptability, and structural transferability, offering a practical and generalizable approach for future time-series forecasting model design.

\begin{table}[!b]
\centering
\caption{Ablation study on the ETTh1 dataset. Lower is better. The best results are in \best{bold} and the second best are \second{underlined}.}
\label{tab:ablation-etth1}
\begin{tabular}{l
                S[table-format=1.3]
                S[table-format=1.3]
                S[table-format=1.3]
                S[table-format=1.3]}
\toprule
\textbf{Variant} 

& \multicolumn{2}{c}{\textbf{ETTh1-96}} 
& \multicolumn{2}{c}{\textbf{ETTh1-192}} \\
& \textbf{MSE} & \textbf{MAE} 
& \textbf{MSE} & \textbf{MAE} \\
\midrule
FM-LLM                   & \best{0.342} & \best{0.380} & \best{0.377} & \best{0.403} \\
w/o FAN                  & 0.351 & 0.390 & 0.381 & 0.411 \\
w/o MOE                  & 0.348 & 0.390 & 0.383 & 0.412 \\
w/o LLM                  & 0.370 & 0.406 & 0.404 & 0.428 \\
w/o $\mathcal{L}_{\text{freq}}$ & 0.352 & 0.387 & 0.389 & 0.412 \\
w/o $\mathcal{L}_{\text{time}}$ & 0.357 & 0.393 & 0.391 & 0.416 \\
FAN $\rightarrow$ Linear & 0.361 & 0.394 & 0.399 & 0.418 \\
\bottomrule
\end{tabular}
\end{table}


\subsection{Ablation study}
To thoroughly assess the contributions of individual modules in FM-LLM, we conducted two sets of ablation studies on the ETT dataset series, as reported in Table~\ref{tab:ablation-etth1}. The study (Table~\ref{tab:ablation-etth1}) examines the effect of removing or replacing major architectural components, including the frequency-aware encoder, the FAN-MoE decoder, the pretrained LLM backbone, and the hybrid loss function. Results across two forecasting horizons reveal that removing the FAN encoder consistently leads to significant performance degradation, especially for longer horizons, highlighting its critical role in capturing periodic structures in the frequency domain. Similarly, the removal of the MoE decoder, which is responsible for adaptive decoding of heterogeneous sequence patterns, results in notable performance drops, particularly on complex multivariate inputs. Under the condition of strict parameter parity, removing the pre-trained large language model weights and training from scratch resulted in the most significant degradation in model performance. This indicates that the performance gains are attributed not only to model capacity but also to the transferred pre-training priors.

To strictly disentangle the contribution of pre-trained linguistic knowledge from the model capacity, we conducted a comprehensive ablation study. Specifically, we aim to verify whether the performance gains of FM-LLM stem from the universal patterns embedded in the LLM backbone or merely from the deep network architecture. The results in Table \ref{tab:pretrain_ablation_full} demonstrate that while the pre-trained backbone offers moderate accuracy gains in full-data settings, its contribution is decisive in few-shot scenarios. Specifically, the scratch-trained model suffers from severe overfitting when data is scarce (e.g., MSE spikes to 0.729 on ETTh1), whereas FM-LLM maintains robust performance (MSE 0.451), confirming that the pre-trained weights are essential for generalization capabilities.
\begin{table}[h]

\caption{Comparisons of Pre-training Effectiveness on ETTh1 and ETTm1 datasets under Full (100\%) and Few-shot (10\%) settings. \textbf{FM-LLM} denotes our proposed method with pre-trained weights. \textbf{w/o Pretrain} denotes the same architecture with random initialization. \textbf{Scratch-trained} denotes a standard Transformer \textbf{of comparable parameter size} trained from scratch. The best results are highlighted in \textbf{bold}.}

\label{tab:pretrain_ablation_full}

\resizebox{\textwidth}{!}{%

\begin{tabular}{c|l|cc|cc|cc|cc}

\toprule

\multirow{2}{*}{\textbf{Dataset}} & \multicolumn{1}{c|}{\multirow{2}{*}{\textbf{Model}}} & \multicolumn{2}{c|}{\textbf{96}} & \multicolumn{2}{c|}{\textbf{192}} & \multicolumn{2}{c|}{\textbf{336}} & \multicolumn{2}{c}{\textbf{720}} \\ \cmidrule(l){3-10} 

 & \multicolumn{1}{c|}{} & MSE & MAE & MSE & MAE & MSE & MAE & MSE & MAE \\ \midrule

\multicolumn{10}{c}{\textit{\textbf{Full Training Data (100\%)}}} \\ \midrule

\multirow{3}{*}{ETTh1} & FM-LLM & \textbf{0.342} & \textbf{0.380} & \textbf{0.377} & \textbf{0.403} & \textbf{0.395} & \textbf{0.415} & \textbf{0.397} & \textbf{0.429} \\

 & w/o Pretrain & 0.348 & 0.387 & 0.381 & 0.409 & 0.398 & 0.422 & 0.408 & 0.441 \\

 & Scratch-trained & 0.370 & 0.406 & 0.404 & 0.428 & 0.420 & 0.440 & 0.432 & 0.460 \\ \midrule

\multirow{3}{*}{ETTm1} & FM-LLM & \textbf{0.276} & \textbf{0.327} & \textbf{0.319} & \textbf{0.356} & \textbf{0.351} & \textbf{0.379} & \textbf{0.406} & \textbf{0.415} \\

 & w/o Pretrain & 0.285 & 0.337 & 0.325 & 0.363 & 0.362 & 0.385 & 0.432 & 0.423 \\

 & Scratch-trained & 0.296 & 0.347 & 0.341 & 0.376 & 0.376 & 0.399 & 0.439 & 0.436 \\ \midrule

\multicolumn{10}{c}{\textit{\textbf{Few-shot Training Data (10\%)}}} \\ \midrule

\multirow{3}{*}{ETTh1} & FM-LLM & \textbf{0.451} & \textbf{0.454} & \textbf{0.481} & \textbf{0.472} & \textbf{0.497} & \textbf{0.486} & \textbf{0.525} & \textbf{0.514} \\

 & w/o Pretrain & 0.476 & 0.466 & 0.505 & 0.481 & 0.529 & 0.496 & 0.580 & 0.527 \\

 & Scratch-trained & 0.551 & 0.511 & 0.595 & 0.537 & 0.620 & 0.554 & 0.696 & 0.597 \\ \midrule

\multirow{3}{*}{ETTm1} & FM-LLM & \textbf{0.383} & \textbf{0.403} & \textbf{0.422} & \textbf{0.427} & \textbf{0.463} & \textbf{0.450} & \textbf{0.561} & \textbf{0.498} \\

 & w/o Pretrain & 0.405 & 0.416 & 0.440 & 0.438 & 0.476 & 0.459 & 0.567 & 0.504 \\

 & Scratch-trained & 0.533 & 0.466 & 0.580 & 0.491 & 0.639 & 0.519 & 0.846 & 0.603 \\ \bottomrule

\end{tabular}%

}

\end{table}


Further, the hybrid time-frequency loss proves to be an integral design choice. Ablating the frequency-domain loss slightly reduces the model’s ability to align periodic behavior, while removing the time-domain loss has a more pronounced negative impact on numerical prediction accuracy. These findings indicate that both loss components are complementary: time-domain loss governs local precision, while frequency-domain loss ensures global structural consistency. Additionally, replacing the FAN module with a linear projection leads to a consistent drop in performance, underscoring the importance of explicit spectral modeling over simple linear mappings.

\begin{figure}[!t]
    \centering
    \includegraphics[width=1.0\linewidth]{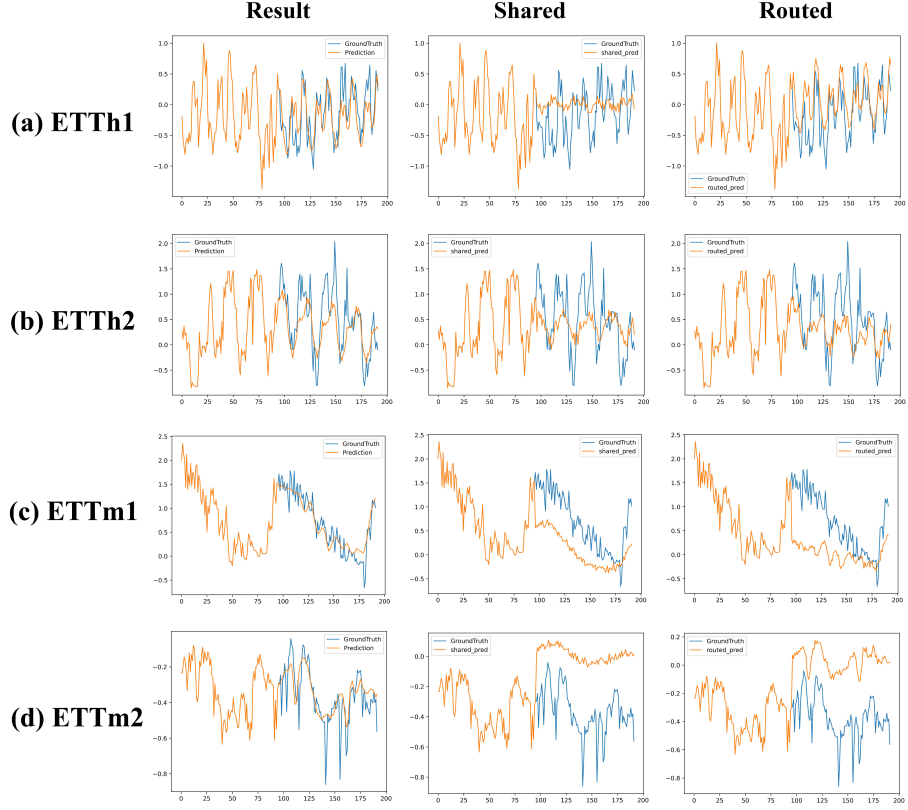}
    \caption{The case study on ETT datasets. For convenience, we set both the lookback length and the forecasting horizon to 96. The blue line represents 
the ground-truth and the orange line denotes the forecasting result.}
    \label{fig:result}
\end{figure}

\subsection{Visualization Analysis}
Figure~\ref{fig:result} presents the prediction composition of the FM-LLM model on four ETT datasets (ETTh1, ETTh2, ETTm1, and ETTm2), visually illustrating the structural division of labor and cooperative mechanisms between the shared Fourier Experts and Routed Experts. According to the model design, the final prediction result (denoted as “Result”) is obtained by a weighted summation of the outputs from both expert groups. As clearly observed in the figure, the two types of experts exhibit distinct modeling behaviors: Fourier Experts tend to learn stable periodic structures and trend dynamics, producing smooth and consistent outputs that demonstrate strong global modeling capabilities. In contrast, Routed Experts exhibit a higher degree of dynamic responsiveness, adapting flexibly to local disturbances, phase shifts, or abrupt pattern changes in the input sequence, and providing structural compensation accordingly.

In the ETTh1 dataset which is characterized by strong periodicity and smooth variations—the output of Fourier Experts closely aligns with the final Result, indicating their dominant role in modeling the overall sequence. The contributions from Routed Experts are limited to subtle adjustments at a few inflection points, offering marginal refinements to the Fourier structure. In ETTh2, while the overall trend is still primarily captured by the Fourier Experts, Routed Experts play a more significant role when phase misalignments occur, providing necessary structural corrections that bring the Result closer to the ground truth. This demonstrates a clear complementary relationship between the two expert groups.

The distinction becomes more pronounced in the medium-frequency minute-level datasets ETTm1 and ETTm2, which contain substantial high-frequency fluctuations and short-period disruptions. In these settings, the outputs of Fourier Experts struggle to fit such non-stationary patterns, often resulting in lagging or overly smoothed predictions. Routed Experts, however, exhibit stronger localized responsiveness, effectively capturing abrupt dynamics in the input series. By integrating the temporal sensitivity of Routed Experts with the periodic backbone provided by Fourier Experts, the final Result curve achieves superior fitting performance compared to either expert group alone. This phenomenon further validates a core design hypothesis of FM-LLM: through the expert division and dynamic routing mechanism in its MoE decoder, the model can automatically identify and activate the most appropriate expert sub-network based on the characteristics of each input subsequence. This enables structured modeling of varying frequencies, fluctuation rates, and heterogeneous patterns. The expert cooperation–division–reconstruction process not only enhances the accuracy of long-sequence forecasting, but also provides a mechanistic explanation for the model’s robust generalization in both few-shot and zero-shot scenarios.
\begin{figure*}[!t]
    \centering
    \includegraphics[width=1\linewidth]{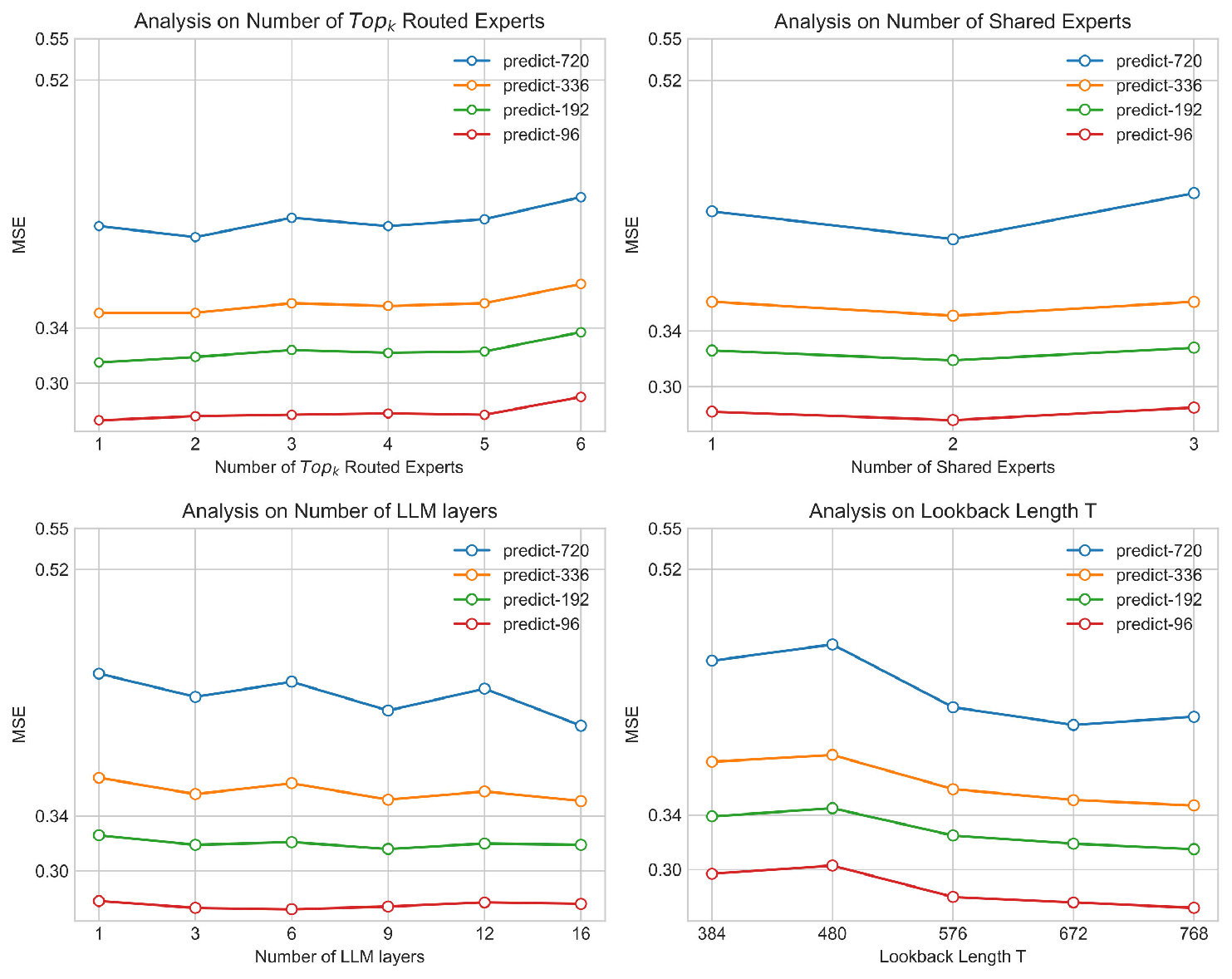}
    \caption{Analysis of hyperparameter sensitivity on ETTm1 dataset.}
    \label{fig:ETTm1-sen}
\end{figure*}

Figure\ref{fig:ETTm1-sen} presents a comprehensive analysis of FM-LLM’s sensitivity to four key architectural and training hyperparameters on the ETTm1 dataset: the number of routed experts (TopK), the number of Fourier experts, the number of LLM layers, and the lookback input length 
$\mathrm{T}$. Each subfigure reports the MSE for different prediction horizons.Based on our analysis, we have made the following observations:

(1) The model exhibits stable performance across different TopK values. However, when the number of routed experts exceeds 4, a slight increase in MSE is observed, likely due to reduced expert specialization and routing redundancy. The best performance is generally achieved with \textit{TopK} = 2 or 3, which balances diversity and computational efficiency. (2) Introducing Fourier experts slightly improves model performance. Using two Fourier experts yields the most consistent gains across prediction lengths. Adding more (e.g., 3) leads to diminishing returns, suggesting that a small Fourier capacity suffices to capture general temporal patterns. (3) FM-LLM is relatively insensitive to the number of backbone transformer layers. While the overall trends are flat, slight improvements are observed around 9--12 layers, particularly for long-horizon prediction settings. This indicates that moderate-depth architectures can enhance representation power without overfitting. (4) Increasing the lookback length consistently reduces MSE across all forecasting horizons. Performance improves significantly from $T = 384$ to $T = 768$, beyond which the benefit saturates. This suggests that $T = 672$ provides an effective historical context for long-term forecasting tasks.
Overall, FM-LLM demonstrates robustness and adaptability to a range of hyperparameter choices, reducing the burden of manual tuning and improving its applicability to real-world time-series forecasting scenarios.
We also investigated the impact of the number of FAN layers on forecasting performance. The detailed experimental results are provided in \hyperref[app5]{Appendix~B} . Empirical evidence suggests that stacking additional layers does not yield performance gains; specifically, the single-layer architecture consistently outperformed the 2-layer and 3-layer configurations across the two datasets. Therefore, we adopt the 1-layer FAN as the default setting for efficiency and robustness
\begin{figure*}[!t]
    \centering
    \includegraphics[width=1\linewidth]{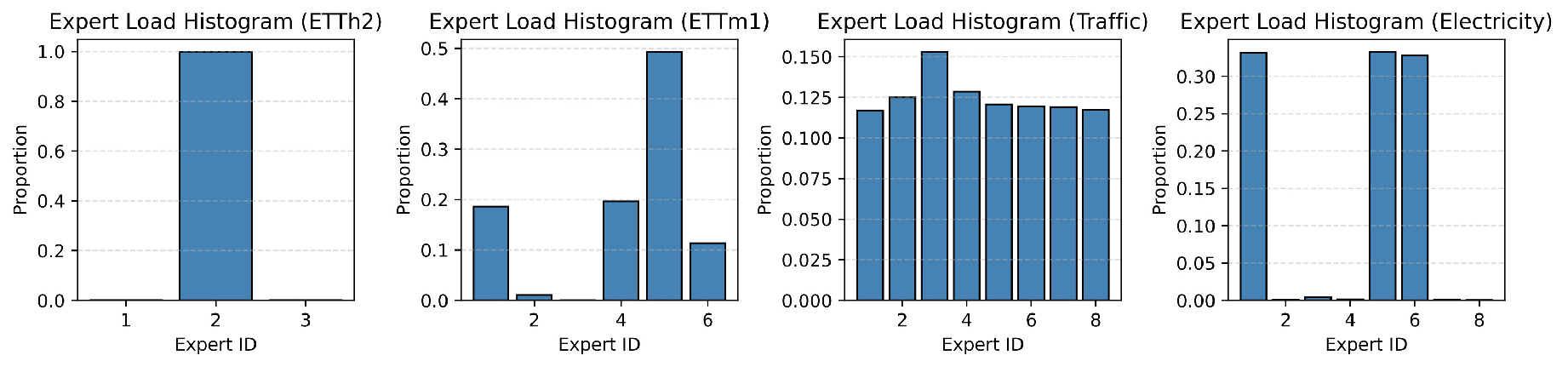}
    \caption{Expert load histograms across different datasets.}
    \label{fig:expert}
\end{figure*}

Figure~\ref{fig:expert} presents the expert load histograms across four representative datasets. On ETTh2, the routing distribution collapses to a single expert, suggesting that the temporal patterns are simple and can be modeled without expert diversity. ETTm1 exhibits moderate imbalance, where one expert dominates while others are partially activated. In contrast, the Traffic dataset demonstrates a near-uniform expert utilization, reflecting its high-dimensional, heterogeneous structure and justifying the necessity of distributed expert computation. For Electricity, the router favors three experts while suppressing others, indicating partial mode specialization but still revealing under-utilization. These patterns confirm that the MoE decoder adapts dynamically to data complexity and structural diversity.

\begin{figure*}[!t]
  \centering
  \begin{subfigure}[b]{0.45\textwidth}
    \includegraphics[width=\textwidth]{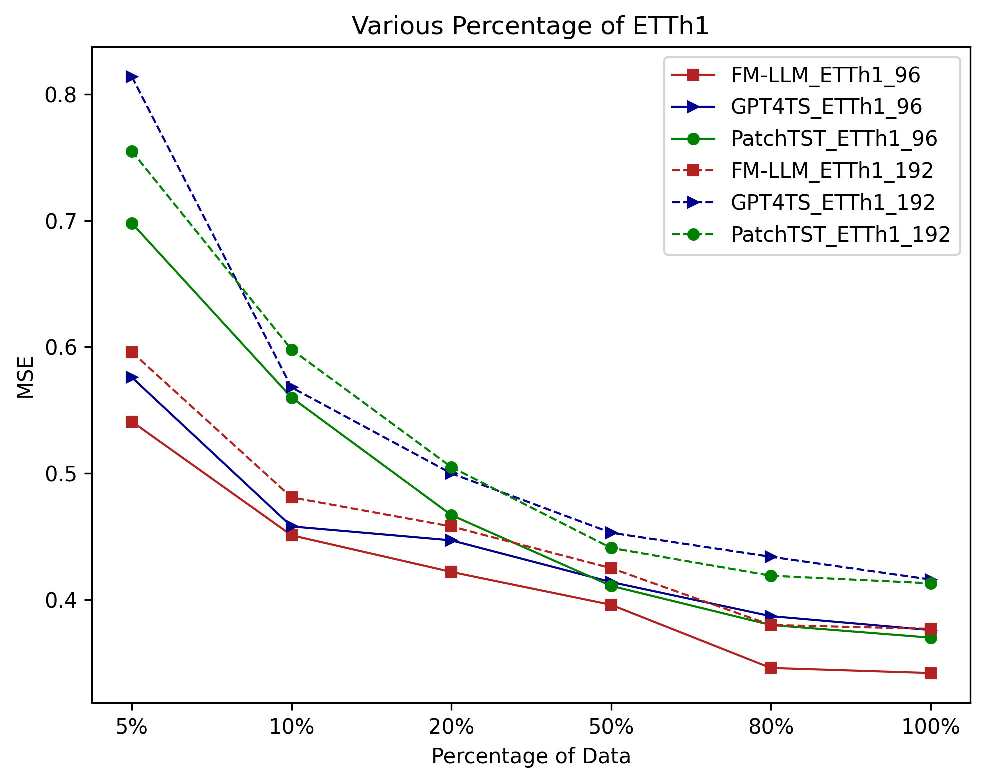}
        \caption{ETTh1}
  \end{subfigure}
  \hfill
    \begin{subfigure}[b]{0.45\textwidth}
    \includegraphics[width=\textwidth]{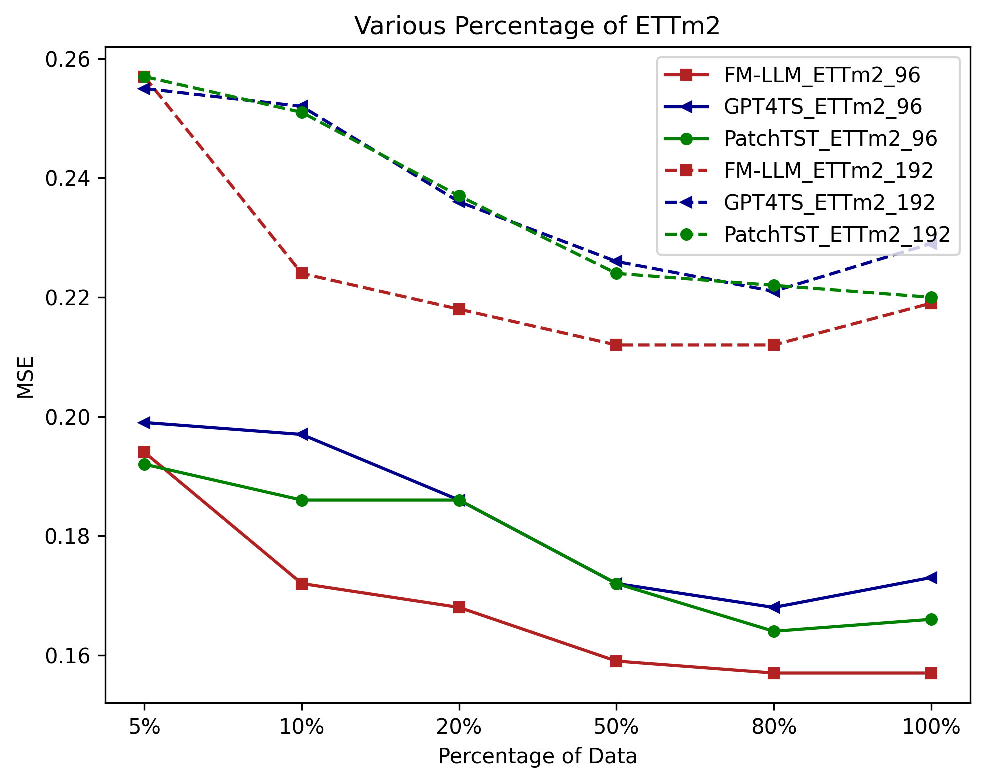}
    \caption{ETTm2}
  \end{subfigure}
  \caption{Results on various percentages of ETTh1 and ETTm2. Line color represents different models and line
style means various prediction lengths $\mathrm{T} \in \{96, 192\}$}
  \label{dataset percent}
\end{figure*}

Figure~\ref{dataset percent} presents model performance under varying training data sizes on the ETTh1 and ETTm2 datasets. FM-LLM consistently outperforms both GPT4TS and PatchTST across all data availability levels and prediction horizons, with especially large margins under low-resource settings. This demonstrates the model's strong few-shot generalization ability, likely attributed to its frequency-aware encoding and expert-based modular decoding design.

Interestingly, we observe a non-monotonic performance trend on ETTm2, where FM-LLM achieves its best accuracy not at 100\% supervision, but rather at 50\% and 80\% training data. This indicates that increasing data volume does not necessarily yield better performance—a phenomenon previously observed in transformer-based models~\cite{xue2024card}. However, we argue that this limitation is not inherent to the transformer architecture itself, but rather stems from distribution shift within the dataset. In time-series benchmarks like ETTm2, earlier and later data segments often differ in seasonal patterns, anomaly frequency, or system conditions. As a result, using the entire dataset for training may introduce inconsistent or noisy patterns, leading to suboptimal generalization. In contrast, subsets like 50\%–80\% may better align with the test distribution, enabling stronger predictive accuracy.

\begin{figure*}[!htbp]
  \centering
  \begin{subfigure}[b]{0.45\textwidth}
    \includegraphics[width=\textwidth]{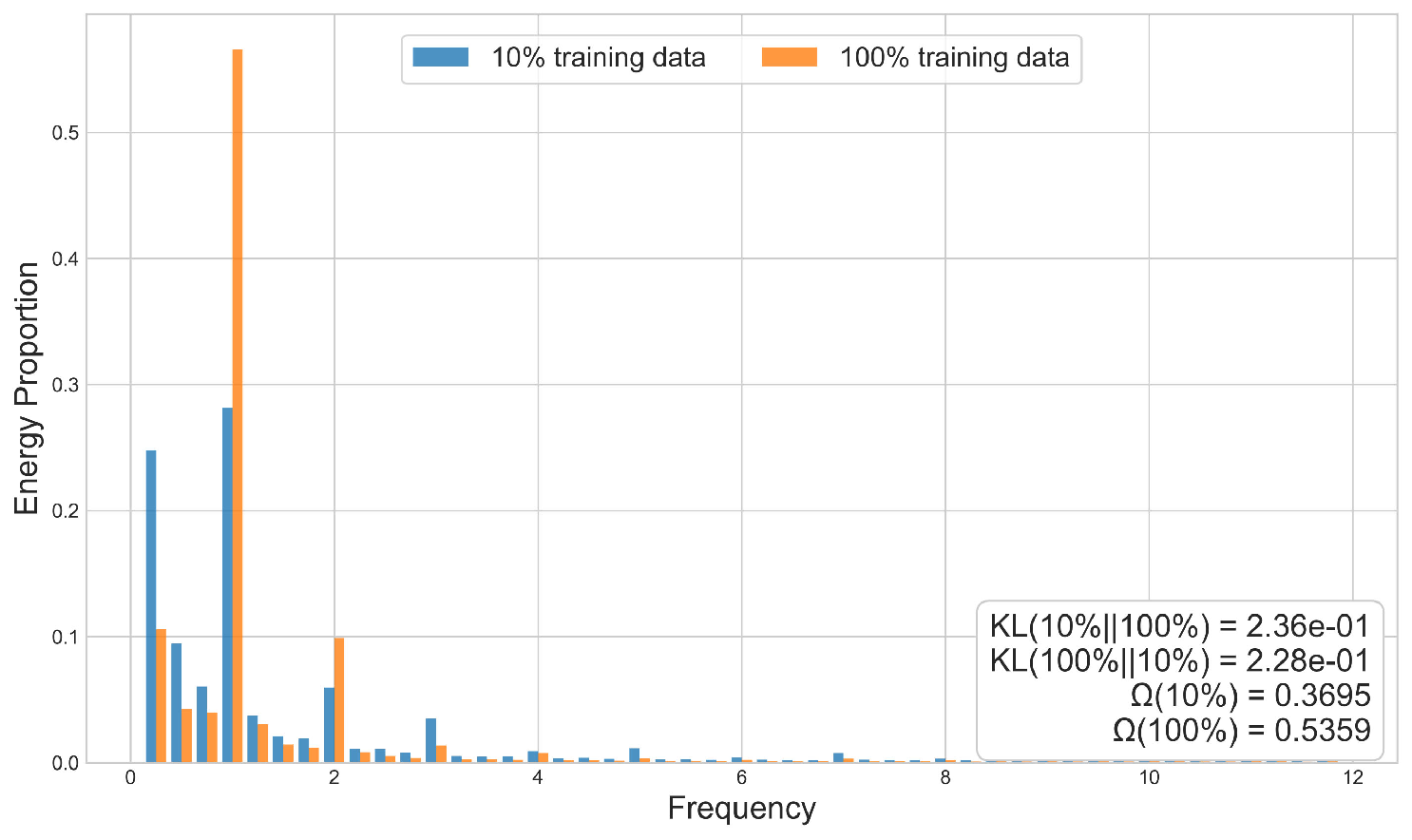}
    \caption{ETTh1}
  \end{subfigure}
  \hfill
  \begin{subfigure}[b]{0.45\textwidth}
    \includegraphics[width=\textwidth]{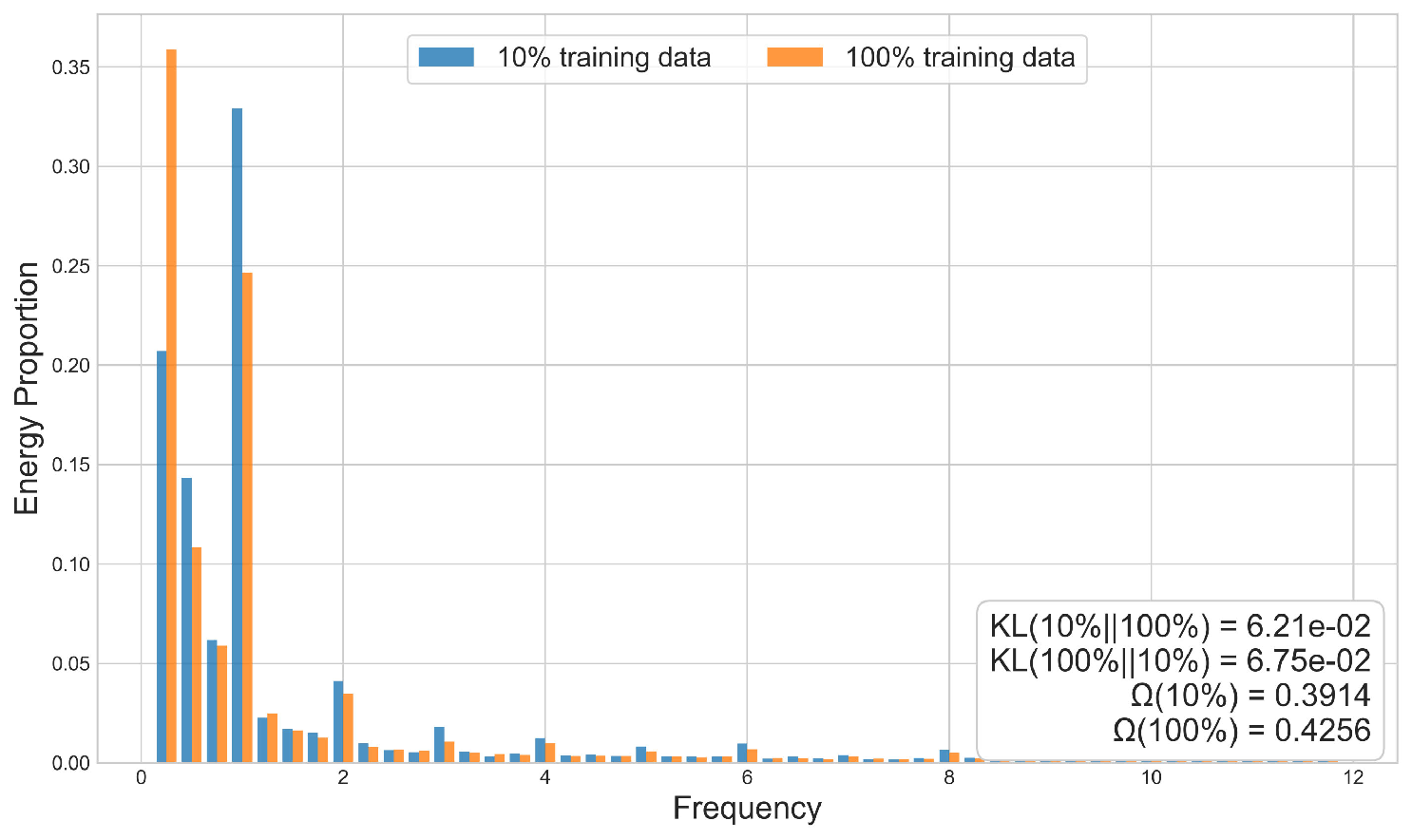}
    \caption{ETTh2}
  \end{subfigure}
  \vspace{0.5em}
  \begin{subfigure}[b]{0.45\textwidth}
    \includegraphics[width=\textwidth]{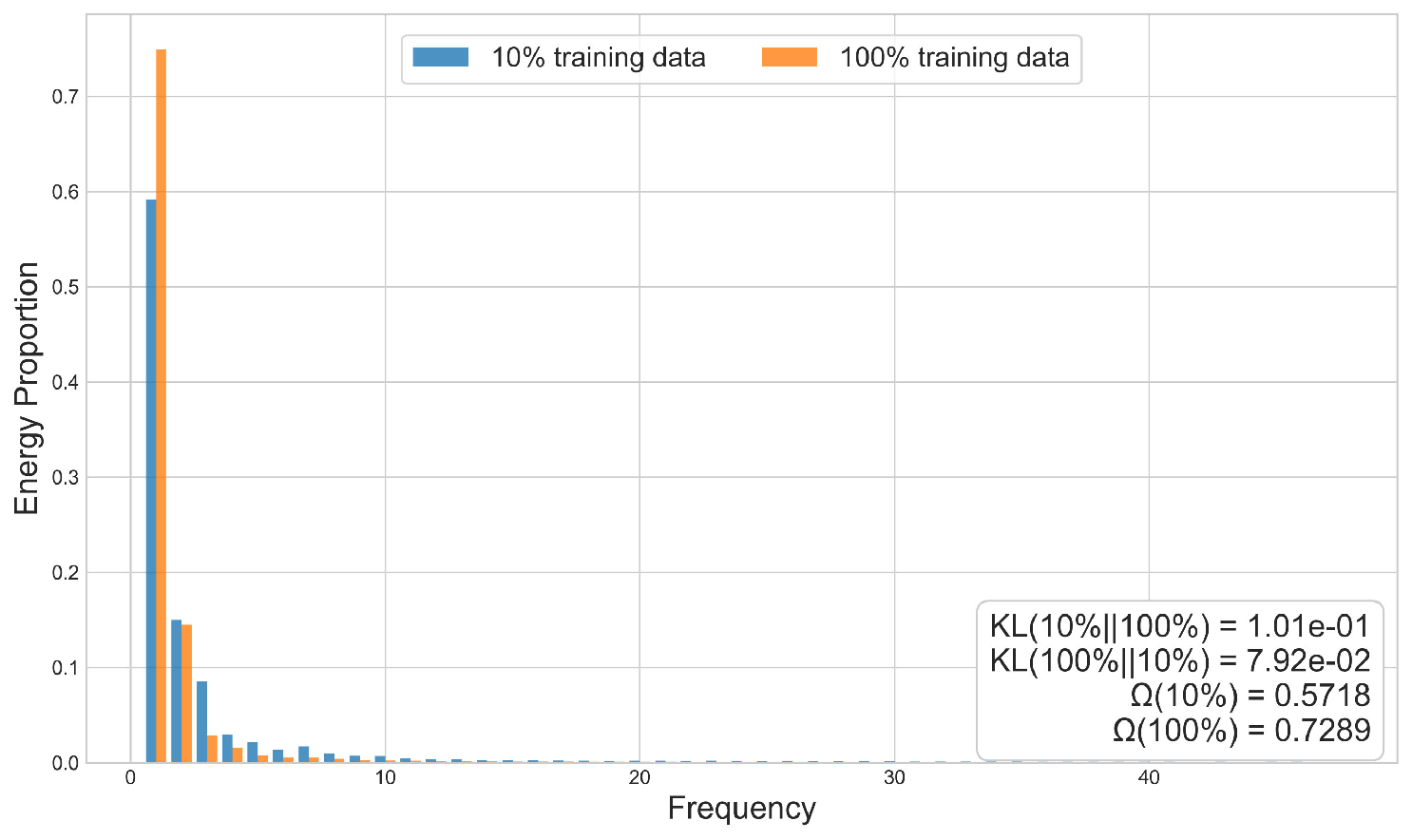}
    \caption{ETTm1}
  \end{subfigure}
  \hfill
    \begin{subfigure}[b]{0.45\textwidth}
    \includegraphics[width=\textwidth]{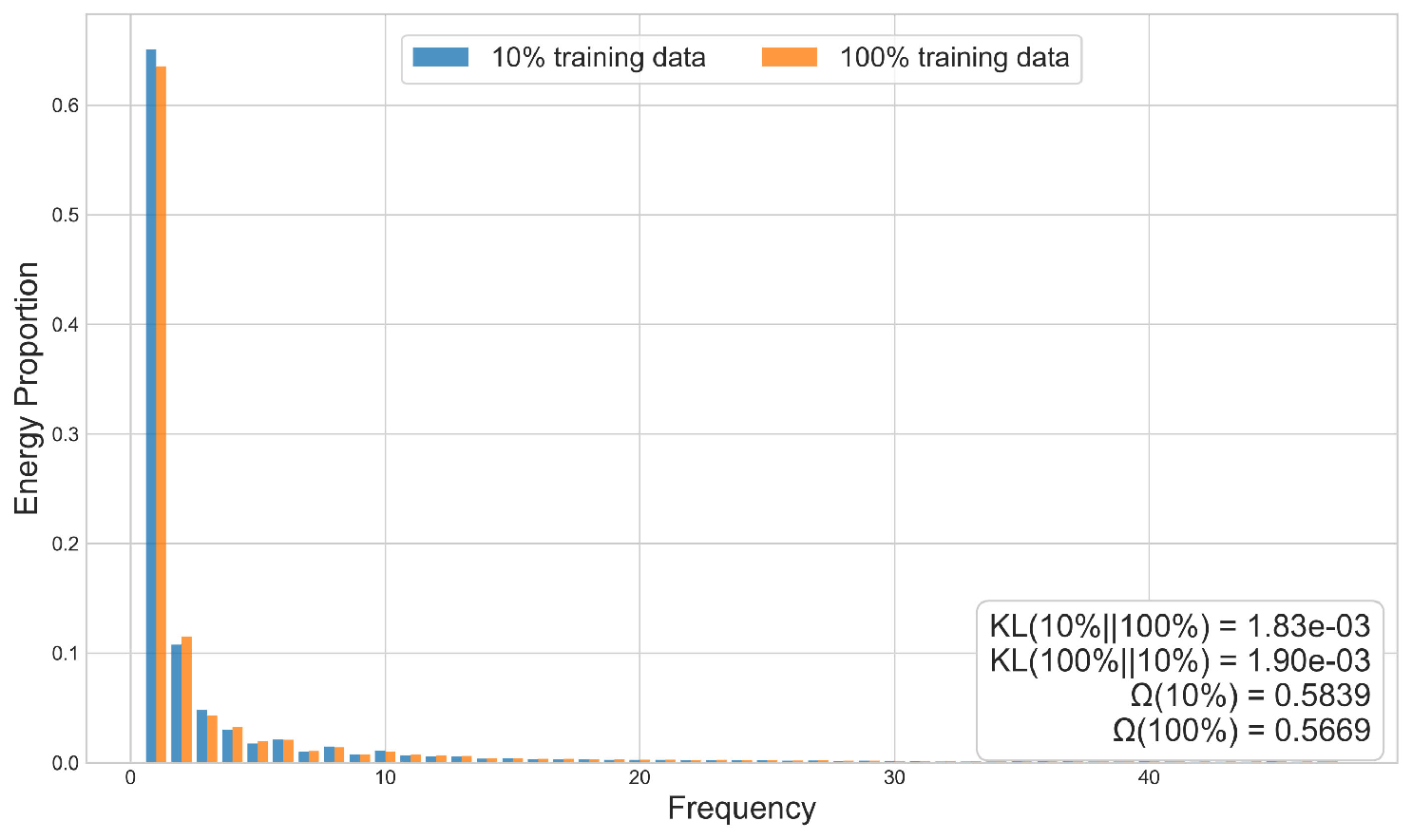}
    \caption{ETTm2}
  \end{subfigure}
  
  \caption{Token-wise frequency energy proportion comparison between 10\% and 100\% training data on the ETT dataset.}
  \label{fig:spectrogram_comparison}
\end{figure*}

\subsection{Frequency-Domain Analysis of Few-Shot Generalization}
Figure~\ref{fig:spectrogram_comparison} provides a frequency–domain explanation for the heterogeneous few–shot performance of \textsc{FM-LLM} across the four ETT benchmarks.  
Using a sliding window of length~$96$, we compute the \emph{averaged} power–spectral density (PSD) for the \mbox{$10\,\%$} and \mbox{$100\,\%$} training subsets, quantify their divergence via the Kullback--Leibler (KL) distance, and assess sequence predictability with the forecastability~\cite{pmlr-v28-goerg13} index
$\Omega = 1 - H/\ln N$.\footnote{$H$ denotes the Shannon entropy of the positive-frequency PSD; $N$ is the number of positive frequency bins.}

\medskip
ETTh1 and ETTm1 exhibit the largest spectral mismatches (KL~=~$2.36\times10^{-1}$ and $1.01\times10^{-1}$, respectively) and a marked drop in~$\Omega$ ($0.536\!\rightarrow\!0.370$;
$0.729\!\rightarrow\!0.572$), indicating that high- and mid-frequency components are poorly represented when only \mbox{$10\,\%$} of the data are available—precisely the datasets where \textsc{FM-LLM} records the highest few-shot error in Table~\ref{tab:example_results}.  
In contrast, ETTm2 shows an almost perfect spectral overlap (KL~=~$1.38\times10^{-3}$) and a virtually unchanged~$\Omega$, implying that even a small subset preserves its dominant frequency structure and thus sustains robust accuracy.

These findings suggest that \textsc{FM-LLM}'s few-shot generalization hinges not only on its frequency-aware architecture but also on whether the reduced training set faithfully reproduces the original energy distribution; spectral divergence therefore emerges as a key prior for estimating few-shot difficulty and expected performance.

\subsection{Analysis of Expert Specialization}
To provide a deeper insight into the internal working mechanism of FM-LLM, we conduct a comprehensive visualization analysis on the Traffic dataset. Specifically, we examine the model from three perspectives: the frequency response of different experts, the specialization patterns of experts, and the temporal dynamics of the routing strategy.

\begin{figure}[H]
    \centering
    \includegraphics[width=0.75\linewidth]{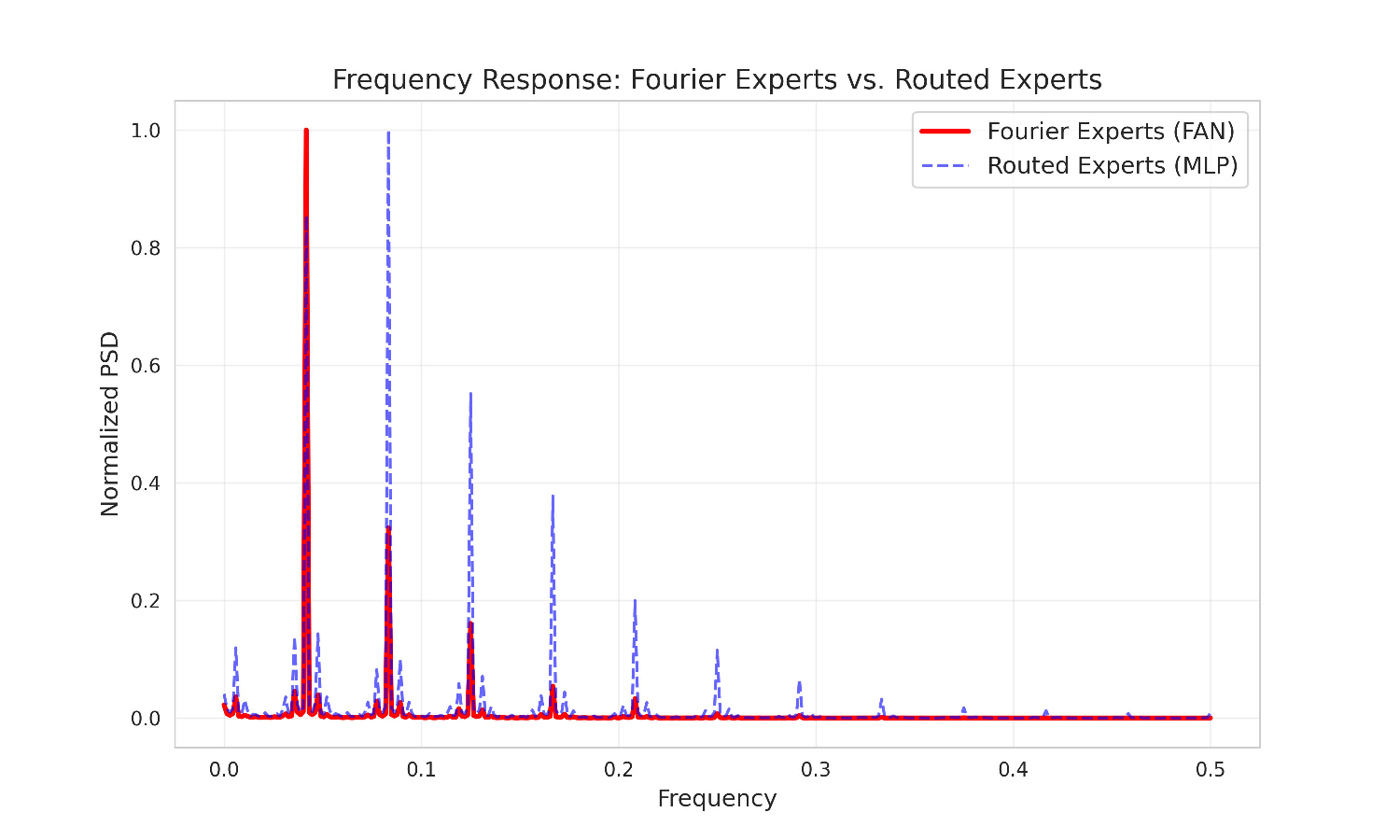}
    \caption{Frequency Response Analysis on Traffic Datasets}
    \label{freq}
\end{figure}

Frequency Domain Decomposition. As illustrated in Figure \ref{freq}, we analyze the spectral characteristics of the learned representations. The frequency response comparison reveals distinct roles: the Fourier experts effectively capture the dominant low-frequency components, which correspond to the strong daily and weekly periodicities inherent in the Traffic dataset. In contrast, the Routed experts exhibit a more diverse spectral focus, attending to high-frequency variations that represent local fluctuations and transient traffic anomalies. This complementary frequency coverage ensures that the model preserves global trends while retaining sensitivity to detailed temporal dynamics.

\begin{figure}[H]
    \centering
    \includegraphics[width=1.0\linewidth]{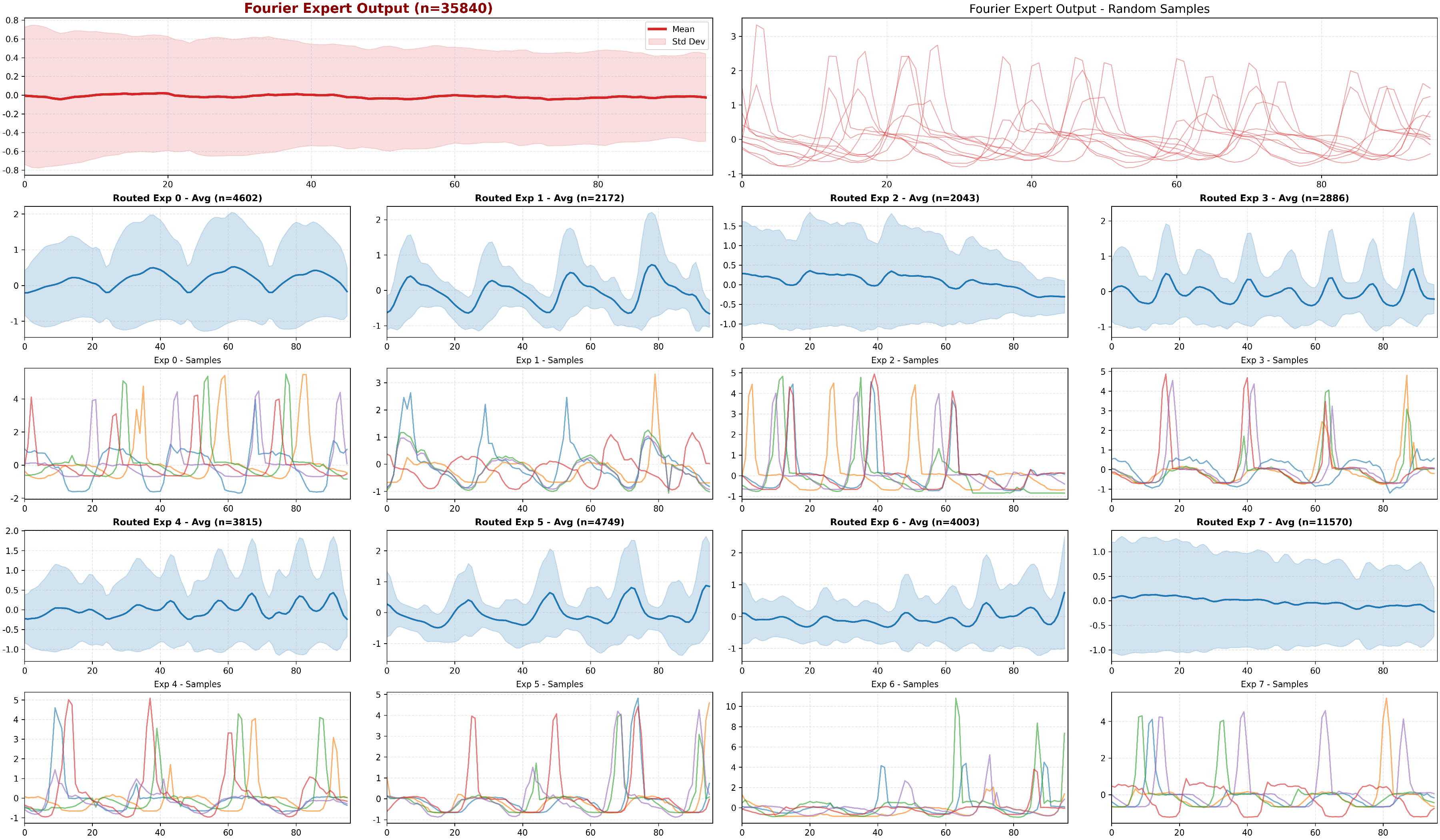}
    \caption{Visualization of Expert Specialization}
    \label{special}
\end{figure}

\begin{figure}[H]
    \centering
    \includegraphics[width=1.0\linewidth]{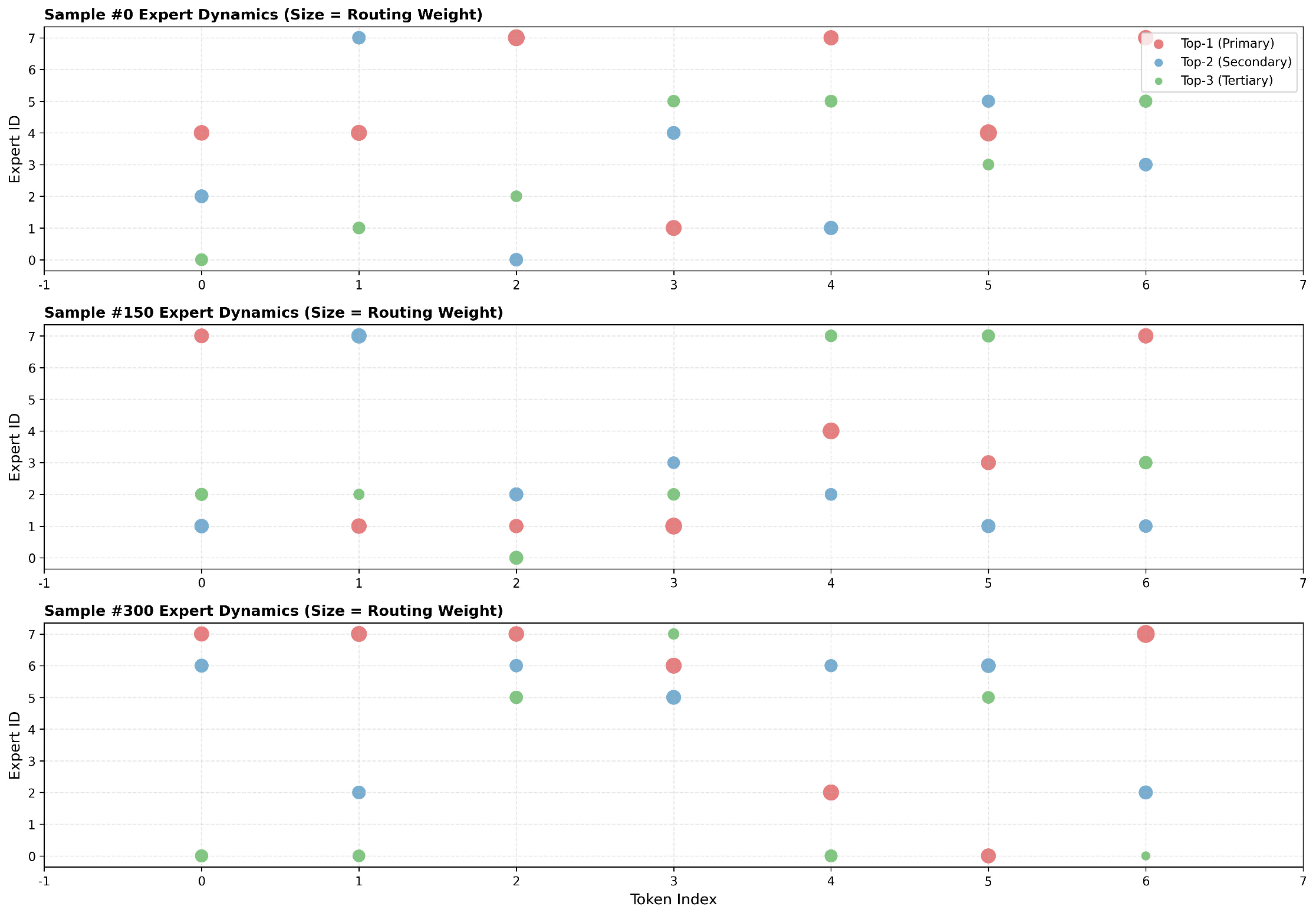}
    \caption{Temporal Dynamics of Top-3 Expert Routing}
    \label{fig:placeholder}
\end{figure}

Expert Specialization Patterns. Figure \ref{special} visualizes the learned preferences of different experts across the traffic data samples. We observe a clear emergence of functional specialization without explicit supervision. Specific experts are consistently activated for samples exhibiting high-volatility congestion patterns, whereas others specialize in modeling stable, free-flow traffic periods. This heterogeneity suggests that the MoE framework successfully decomposes the complex traffic distribution into distinct sub-regimes, allowing each expert to master a specific subset of traffic behaviors.

Temporal Routing Dynamics. To understand how the model adapts over time, Figure \ref{fig:placeholder} depicts the temporal evolution of the Top-3 activated experts over a continuous forecasting window. The routing mechanism demonstrates dynamic adaptability; This indicates that the router dynamically modulates the combination of experts in response to the changing statistical properties of the traffic flow, effectively switching strategies to match the instantaneous context.

\subsection{Efficiency Analysis}

To provide a thorough efficiency assessment, we compare FM-LLM with several competitive baselines,
including LLM-based forecasting methods (GPT4TS and TimeLLM) and strong non-LLM backbones (PatchTST and DLinear).
All experiments are implemented in PyTorch with a fixed batch size of 128.
We report four efficiency metrics: the number of trainable parameters, inference latency (ms/iter),
GPU memory footprint, and peak GPU memory usage. Importantly, all memory statistics are measured
\emph{during inference on a trained model} (forward pass only), without gradient computation.
Moreover, FM-LLM is trained \emph{once} and can be directly applied to \emph{multiple} prediction lengths
without retraining, which further improves its practicality in real-world deployments.

As shown in Table~\ref{tab:traffic-efficiency}, FM-LLM exhibits a favorable accuracy--efficiency trade-off on the Traffic dataset.
Under the long-horizon setting (Pred=720), FM-LLM uses 14.3M trainable parameters and runs at 143.48 ms/iter,
whereas TimeLLM is an order of magnitude slower (1056.0 ms/iter) and requires substantially more GPU memory
(22.6 GB footprint and 11.1 GB peak). Compared with PatchTST, FM-LLM incurs higher computation and memory cost,
yet it remains markedly more efficient than LLM-based baselines. Under the short-horizon setting (Pred=96),
FM-LLM further reduces latency to 17.93 ms/iter while maintaining a stable memory footprint (3.23 GB),
whereas TimeLLM still suffers from extremely high latency (1039.6 ms/iter) and memory usage (22.6 GB footprint).
Overall, FM-LLM substantially improves practicality over existing LLM-based forecasting baselines
while maintaining competitive forecasting performance.

\begin{table}[!b]
\centering
\caption{Efficiency comparison on \textbf{Traffic}. We report trainable parameters, inference time (ms/iter), GPU memory footprint (MB), peak memory (MB), and MSE performance under different prediction lengths. \textit{Note: The higher latency compared to PatchTST is a strategic trade-off for gaining LLM-based few-shot generalization.}}
\label{tab:traffic-efficiency}

\scriptsize
\setlength{\tabcolsep}{3.2pt} 
\renewcommand{\arraystretch}{1.2}

\begin{tabular*}{\textwidth}{@{\extracolsep{\fill}} c|c|c|c|c|c|c|c } 
\toprule
\textbf{Dataset} & \textbf{Pred} & \textbf{Model} & \textbf{MSE} & 
\makecell{\textbf{Trainable}\\\textbf{Params}} & 
\makecell{\textbf{Inference}\\\textbf{Time (ms/it)}} & 
\makecell{\textbf{GPU Mem}\\\textbf{Footprint}} & 
\makecell{\textbf{Peak}\\\textbf{Mem}} \\
\midrule

\multirow{10}{*}{Traffic} & 
\multirow{5}{*}{720} & 
FM-LLM   & \best{0.419}  & 14,329,800  & 143.48  & 3,232   & 3,184  \\
& & GPT4TS  & 0.450 & 94,407,376  & 139.95  & 3,734   & 3,496  \\
& & TimeLLM & 0.430 & 131,312,824 & 1,056.0 & 22,592  & 11,122 \\
& & PatchTST & 0.432 & 6,310,224   & 2.3     & 566     & 205    \\
& & DLinear & 0.466 & 738,720     & 0.47    & 48      & 28     \\
\cmidrule(r){2-8} 

& \multirow{5}{*}{96} & 
FM-LLM   & \best{0.345}  & 14,329,800  & 17.93   & 3,232   & 3,184.8  \\
& & GPT4TS  & 0.388 & 12,617,824  & 138.70  & 3,422   & 3,184    \\
& & TimeLLM & 0.362 & 130,034,248 & 1,039.6 & 22,592  & 11,110.5  \\
& & PatchTST & 0.360 & 1,197,792   & 2.2     & 166.2   & 464      \\
& & DLinear & 0.388 & 98,496      & 0.33    & 28      & 20.3     \\
\bottomrule
\end{tabular*}
\end{table}

\subsection{Spectral Analysis of Model Predictions}

\begin{figure}[!t]
    \centering
    \includegraphics[width=\textwidth]{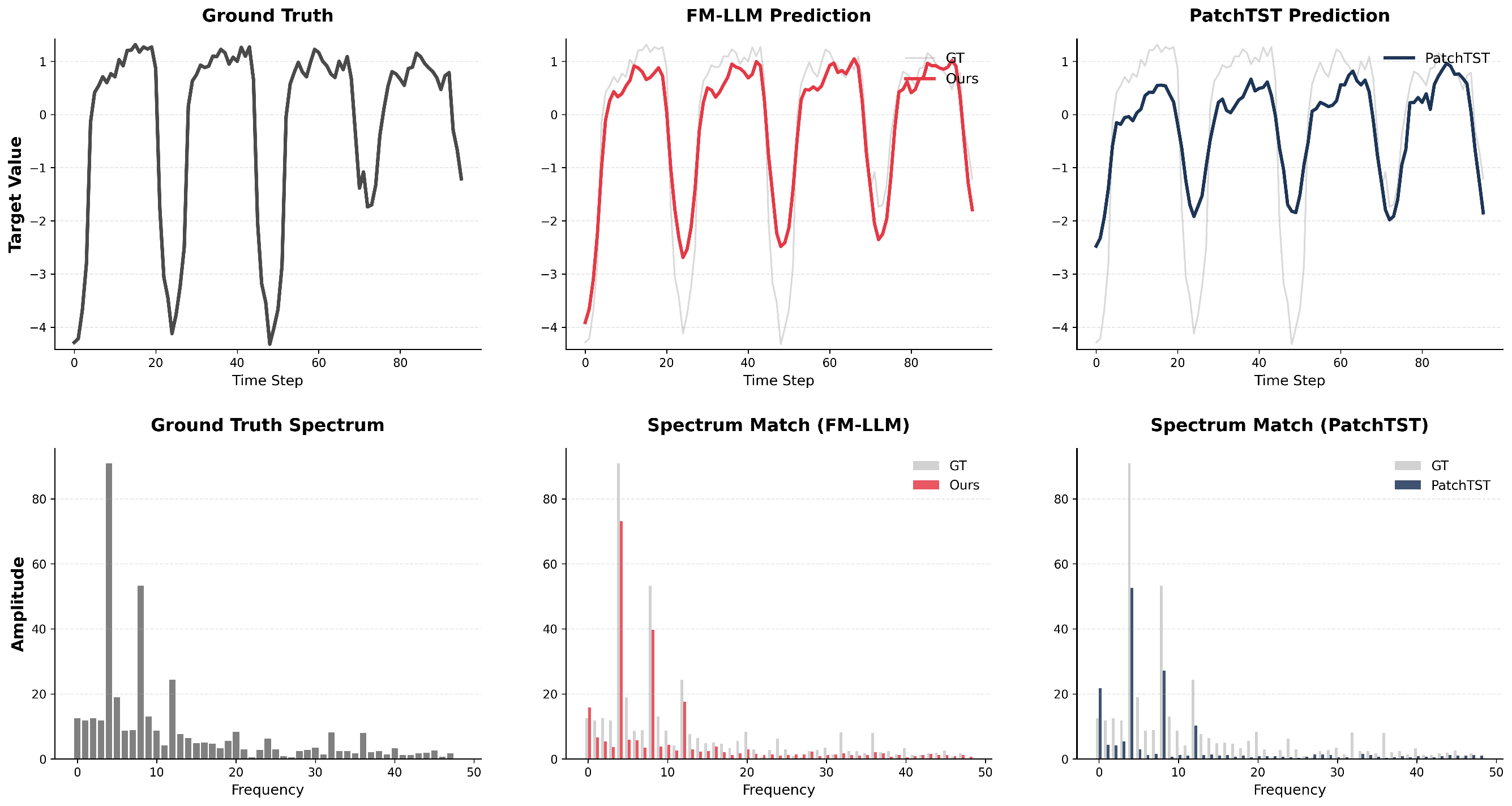}
    \caption{Prediction spectrogram cases from ETTh1 by ground truth and different models under the
input-672-predict-96 settings.}
    \label{fig:appH_spectrum_case}
\end{figure}

To demonstrate the effectiveness of FM-LLM, we visualize the time-domain trajectories and the corresponding frequency spectra of the ground truth and model predictions. As shown in Fig.~\ref{fig:appH_spectrum_case}, we present representative spectrogram cases on ETTh1 under the input-672-predict-96 setting, comparing the ground truth and different models. We observe that FM-LLM aligns better with the dominant frequency components and preserves the harmonic structure more faithfully, which is consistent with its frequency-enhanced token adaptation design.

\section{Conclusion}

In this work, we propose \textbf{FM-LLM}, a frequency-aware framework specifically designed for adapting frozen LLMs to long-term time-series forecasting. Departing from the conventional reliance on prompt engineering, FM-LLM resolves the modality gap and the superimposed entanglement of temporal dynamics through a purposeful, constrained asymmetric coupling. By integrating explicit spectral token alignment at the encoding stage and an asymmetric expert role division at the decoding stage, FM-LLM achieves a structural decoupling of global periodic backbones and local non-stationary residuals. Extensive experiments demonstrate that FM-LLM not only achieves state-of-the-art accuracy across full-data and few-shot settings but also maintains high computational efficiency by avoiding sequence-length expansion. These findings suggest that the structural constraints and spectral inductive biases in FM-LLM provide a robust, interpretable, and efficient solution for cross-modal time-series modeling.

In future work, we aim to explore more lightweight expert modules and parameter-efficient fine-tuning (PEFT) methods, such as LoRA, to further minimize adaptation overhead and improve efficiency. We also plan to develop more robust expert routing mechanisms to stabilize training and ensure balanced expert utilization. Furthermore, we will extend FM-LLM to broader time-series applications, specifically targeting communication network traffic prediction. Given the model's strength in decoupling periodic tidal patterns from bursty anomalies, it holds significant potential for optimizing 5G/6G network slicing and base station load balancing. Finally, at present, the loss function adopts fixed weighting coefficients; in future work, we aim to develop a loss formulation that dynamically adjusts task weights based on gradient magnitudes. Alongside this, we will investigate advanced autoregressive training strategies—such as continuous-variable classification and frequency-domain masked mechanisms—to further enhance sequence modeling performance without compromising spectral continuity.

\clearpage

\appendix

\label{app1}
\section{Full Results on Long-Term and Short-Term Forecasting}

This appendix reports the complete results that are omitted from the main text for clarity.
Table~\ref{tab:m4-full-breakdown} provides the full breakdown of M4 short-term forecasting results by sampling intervals.
Table~\ref{tab:benchmark-results-full} reports long-term forecasting results across prediction lengths \{96, 192, 336, 720\}.

\begin{table}[H]
\centering

\begin{minipage}{\textwidth}
\centering
\captionof{table}{Short-term forecasting results on the M4 benchmark (SMAPE/MASE/OWA). The best results are highlighted in \best{bold red} and the second best are \second{underlined blue}.}
\label{tab:m4-full-breakdown}

{\scriptsize
\renewcommand{\arraystretch}{1.35}
\setlength{\tabcolsep}{5pt}

\resizebox{\textwidth}{!}{
\begin{tabular}{cc*{11}{c}}
\toprule
\multicolumn{2}{c|}{\textbf{Methods}}
& \makecell{\textbf{FM-LLM}\\\textbf{(Ours)}}
& \makecell{\textbf{TIME-LLM}}
& \makecell{\textbf{GPT4TS}}
& \makecell{\textbf{TimesNet}}
& \makecell{\textbf{PatchTST}}
& \makecell{\textbf{N-HiTS}}
& \makecell{\textbf{N-BEATS}}
& \makecell{\textbf{DLinear}}
& \makecell{\textbf{FEDformer}}
& \makecell{\textbf{Stationary}}
& \makecell{\textbf{Reformer}}
\\
\midrule

\multicolumn{1}{c|}{\multirow{3}{*}{\rotatebox[origin=c]{90}{\textbf{Yearly}}}}
& \multicolumn{1}{l|}{\textbf{SMAPE}}
& \bestx{13.403} & \secondx{13.419} & 15.11 & 15.378 & 13.477 & 13.422 & 13.487 & 16.965 & 14.021 & 13.717 & 16.169 \\
\multicolumn{1}{c|}{}
& \multicolumn{1}{l|}{\textbf{MASE}}
& \bestx{2.993} & \secondx{3.005} & 3.565 & 3.554 & 3.019 & 3.056 & 3.036 & 4.283 & 3.036 & 3.078 & 3.800 \\
\multicolumn{1}{c|}{}
& \multicolumn{1}{l|}{\textbf{OWA}}
& \bestx{0.787} & \secondx{0.789} & 0.911 & 0.918 & 0.792 & 0.795 & 0.795 & 1.058 & 0.811 & 0.807 & 0.973 \\
\midrule

\multicolumn{1}{c|}{\multirow{3}{*}{\rotatebox[origin=c]{90}{\textbf{Quarterly}}}}
& \multicolumn{1}{l|}{\textbf{SMAPE}}
& \bestx{10.076} & \secondx{10.110} & 10.597 & 10.465 & 10.38 & 10.185 & 10.564 & 12.145 & 11.1 & 10.958 & 13.313 \\
\multicolumn{1}{c|}{}
& \multicolumn{1}{l|}{\textbf{MASE}}
& \bestx{1.177} & \secondx{1.178} & 1.253 & 1.227 & 1.233 & 1.18 & 1.252 & 1.520 & 1.35 & 1.325 & 1.775 \\
\multicolumn{1}{c|}{}
& \multicolumn{1}{l|}{\textbf{OWA}}
& \bestx{0.887} & \secondx{0.889} & 0.938 & 0.923 & 0.921 & 0.893 & 0.936 & 1.106 & 0.996 & 0.981 & 1.252 \\
\midrule

\multicolumn{1}{c|}{\multirow{3}{*}{\rotatebox[origin=c]{90}{\textbf{Monthly}}}}
& \multicolumn{1}{l|}{\textbf{SMAPE}}
& \bestx{12.699} & 12.980 & 13.258 & 13.513 & \second{12.959} & 13.059 & 13.089 & 13.514 & 14.403 & 13.917 & 20.128 \\
\multicolumn{1}{c|}{}
& \multicolumn{1}{l|}{\textbf{MASE}}
& \bestx{0.939} & \secondx{0.963} & 1.003 & 1.039 & 0.97 & 1.013 & 0.996 & 1.037 & 1.147 & 1.097 & 2.614 \\
\multicolumn{1}{c|}{}
& \multicolumn{1}{l|}{\textbf{OWA}}
& \bestx{0.882} & \secondx{0.903} & 0.931 & 0.957 & 0.905 & 0.929 & 0.922 & 0.956 & 1.038 & 0.998 & 1.927 \\
\midrule

\multicolumn{1}{c|}{\multirow{3}{*}{\rotatebox[origin=c]{90}{\textbf{Others}}}}
& \multicolumn{1}{l|}{\textbf{SMAPE}}
& 4.861 & \secondx{4.795} & 6.124 & 6.913 & 4.952 & \bestx{4.711} & 6.599 & 6.709 & 7.148 & 6.302 & 32.491 \\
\multicolumn{1}{c|}{}
& \multicolumn{1}{l|}{\textbf{MASE}}
& 3.261 & \secondx{3.178} & 4.116 & 4.507 & 3.347 & \bestx{3.054} & 4.43 & 4.953 & 4.041 & 4.064 & 33.355 \\
\multicolumn{1}{c|}{}
& \multicolumn{1}{l|}{\textbf{OWA}}
& 1.026 & \secondx{1.006} & 1.259 & 1.438 & 1.049 & \bestx{0.977} & 1.393 & 1.487 & 1.389 & 1.304 & 8.679 \\
\midrule

\multicolumn{1}{c|}{\multirow{3}{*}{\rotatebox[origin=c]{90}{\textbf{Average}}}}
& \multicolumn{1}{l|}{\textbf{SMAPE}}
& \bestx{11.840} & \secondx{11.983} & 12.69 & 12.88 & 12.059 & 12.035 & 12.25 & 13.639 & 13.16 & 12.780 & 18.200 \\
\multicolumn{1}{c|}{}
& \multicolumn{1}{l|}{\textbf{MASE}}
& \bestx{1.585} & \secondx{1.595} & 1.808 & 1.836 & 1.623 & 1.625 & 1.698 & 2.095 & 1.775 & 1.756 & 4.223 \\
\multicolumn{1}{c|}{}
& \multicolumn{1}{l|}{\textbf{OWA}}
& \bestx{0.851} & \secondx{0.859} & 0.94 & 0.955 & 0.869 & 0.869 & 0.896 & 1.051 & 0.949 & 0.930 & 1.775 \\
\bottomrule
\end{tabular}}
}
\end{minipage}

\vspace{3pt}

\begin{minipage}{\textwidth}
\centering
\captionof{table}{Forecasting performance (MSE/MAE) on benchmark datasets. AutoTimes* denotes the use of Llama3.2-1B backbone. The best results are highlighted in \best{bold red} and the second best are \second{underlined blue}.}
\label{tab:benchmark-results-full}

{\scriptsize
\renewcommand{\arraystretch}{1.08}
\setlength{\tabcolsep}{2.6pt}

\begin{adjustbox}{max width=\textwidth}
\begin{tabular}{ll*{22}{c}}
\toprule
\multirow{2}{*}{\textbf{Dataset}} & \multirow{2}{*}{\textbf{Models}}
& \multicolumn{2}{c}{\textbf{FM-LLM}}
& \multicolumn{2}{c}{\textbf{AutoTimes*}~\cite{liu2024autotimes}}
& \multicolumn{2}{c}{\textbf{GPT4TS}~\cite{zhou2023one}}
& \multicolumn{2}{c}{\textbf{PRADA}~\cite{LIU2025113478}}
& \multicolumn{2}{c}{\textbf{PatchTST}~\cite{Yuqietal-2023-PatchTST}}
& \multicolumn{2}{c}{\textbf{iTransformer}~\cite{liu2023itransformer}}
& \multicolumn{2}{c}{\textbf{MICN}~\cite{wang2023micn}}
& \multicolumn{2}{c}{\textbf{TimesNet}~\cite{wu2023timesnet}}
& \multicolumn{2}{c}{\textbf{FEDformer}~\cite{zhou2022fedformer}}
& \multicolumn{2}{c}{\textbf{DLinear}~\cite{Zeng_Chen_Zhang_Xu_2023}}
& \multicolumn{2}{c}{\textbf{AMD}~\cite{hu2025adaptive}}
\\
\cmidrule(r){3-4} \cmidrule(r){5-6} \cmidrule(r){7-8} \cmidrule(r){9-10} \cmidrule(r){11-12}
\cmidrule(r){13-14} \cmidrule(r){15-16} \cmidrule(r){17-18} \cmidrule(r){19-20} \cmidrule(r){21-22} \cmidrule(r){23-24}
& & \textbf{MSE} & \textbf{MAE}
& \textbf{MSE} & \textbf{MAE}
& \textbf{MSE} & \textbf{MAE}
& \textbf{MSE} & \textbf{MAE}
& \textbf{MSE} & \textbf{MAE}
& \textbf{MSE} & \textbf{MAE}
& \textbf{MSE} & \textbf{MAE}
& \textbf{MSE} & \textbf{MAE}
& \textbf{MSE} & \textbf{MAE}
& \textbf{MSE} & \textbf{MAE}
& \textbf{MSE} & \textbf{MAE}
\\
\midrule

\multirow{5}{*}{ETTm1}
& 96  & \bestx{0.276} & \bestx{0.327} & 0.289 & 0.344 & 0.292 & 0.346 & 0.288 & 0.342 & 0.290 & 0.342 & 0.309 & 0.365 & 0.316 & 0.364 & 0.338 & 0.375 & 0.764 & 0.416 & 0.299 & 0.343 & \secondx{0.284} & \secondx{0.339}\\
& 192 & \bestx{0.319} & \bestx{0.356} & 0.336 & 0.373 & 0.332 & 0.372 & 0.326 & 0.366 & 0.332 & 0.369 & 0.348 & 0.387 & 0.363 & 0.390 & 0.371 & 0.387 & 0.426 & 0.441 & 0.355 & 0.365 & \secondx{0.322} & \secondx{0.362}\\
& 336 & \secondx{0.351} & \bestx{0.379} & 0.372 & 0.395 & 0.366 & 0.394 & \bestx{0.350} & \secondx{0.384} & 0.366 & 0.392 & 0.378 & 0.406 & 0.408 & 0.426 & 0.410 & 0.411 & 0.445 & 0.459 & 0.369 & 0.386 & 0.360 & 0.380\\
& 720 & \secondx{0.406} & \secondx{0.415} & 0.427 & 0.430 & 0.417 & 0.421 & \bestx{0.399} & \bestx{0.413} & 0.420 & 0.424 & 0.456 & 0.455 & 0.459 & 0.464 & 0.478 & 0.450 & 0.543 & 0.490 & 0.425 & 0.421 & 0.416 & 0.414\\
& avg & \bestx{0.338} & \bestx{0.369} & 0.356 & 0.386 & 0.352 & 0.383 & \secondx{0.341} & 0.376 & 0.351 & 0.381 & 0.373 & 0.403 & 0.387 & 0.411 & 0.400 & 0.406 & 0.448 & 0.452 & 0.362 & 0.379 & 0.347 & \secondx{0.374}\\
\midrule

\multirow{5}{*}{ETTm2}
& 96  & \bestx{0.157} & \bestx{0.239} & 0.183 & 0.266 & 0.173 & 0.262 & \secondx{0.162} & \secondx{0.249} & 0.165 & 0.255 & 0.181 & 0.274 & 0.179 & 0.275 & 0.187 & 0.267 & 0.203 & 0.287 & 0.167 & 0.260 & 0.167 & 0.258\\
& 192 & \secondx{0.219} & \bestx{0.281} & 0.246 & 0.307 & 0.229 & 0.301 & \bestx{0.216} & \secondx{0.288} & 0.220 & 0.292 & 0.243 & 0.316 & 0.262 & 0.326 & 0.249 & 0.309 & 0.269 & 0.328 & 0.224 & 0.303 & 0.221 & 0.294\\
& 336 & 0.275 & \bestx{0.318} & 0.298 & 0.341 & 0.286 & 0.341 & \bestx{0.270} & \bestx{0.324} & 0.274 & 0.329 & 0.297 & 0.351 & 0.305 & 0.353 & 0.321 & 0.351 & 0.325 & 0.366 & 0.281 & 0.342 & \bestx{0.270} & \secondx{0.327}\\
& 720 & \secondx{0.359} & \bestx{0.375} & 0.373 & 0.390 & 0.378 & 0.401 & \bestx{0.350} & \bestx{0.375} & 0.362 & 0.385 & 0.375 & 0.401 & 0.389 & 0.407 & 0.497 & 0.403 & 0.421 & 0.415 & 0.397 & 0.421 & 0.356 & 0.382\\
& avg & \secondx{0.253} & \bestx{0.303} & 0.275 & 0.326 & 0.266 & 0.326 & \bestx{0.250} & \secondx{0.309} & 0.255 & 0.315 & 0.274 & 0.336 & 0.284 & 0.340 & 0.291 & 0.333 & 0.305 & 0.349 & 0.256 & 0.331 & 0.254 & 0.315\\
\midrule

\multirow{5}{*}{ETTh1}
& 96  & \bestx{0.342} & \bestx{0.380} & \secondx{0.358} & 0.398 & 0.376 & 0.397 & 0.361 & \secondx{0.392} & 0.370 & 0.399 & 0.386 & 0.405 & 0.398 & 0.427 & 0.384 & 0.402 & 0.376 & 0.419 & 0.375 & 0.399 & 0.369 & 0.397\\
& 192 & \bestx{0.377} & \bestx{0.403} & \secondx{0.387} & 0.418 & 0.416 & 0.418 & 0.393 & \secondx{0.412} & 0.413 & 0.421 & 0.422 & 0.439 & 0.430 & 0.453 & 0.436 & 0.429 & 0.420 & 0.448 & 0.405 & 0.416 & 0.401 & 0.416\\
& 336 & \bestx{0.395} & \bestx{0.415} & \secondx{0.401} & 0.429 & 0.442 & 0.433 & 0.412 & \secondx{0.424} & 0.422 & 0.436 & 0.444 & 0.457 & 0.440 & 0.460 & 0.491 & 0.469 & 0.459 & 0.465 & 0.439 & 0.443 & 0.418 & 0.427\\
& 720 & \bestx{0.397} & \bestx{0.429} & \secondx{0.425} & 0.450 & 0.477 & 0.456 & 0.442 & 0.458 & 0.447 & 0.466 & 0.500 & 0.498 & 0.491 & 0.509 & 0.521 & 0.500 & 0.506 & 0.507 & 0.472 & 0.490 & 0.439 & 0.454\\
& avg & \bestx{0.378} & \bestx{0.407} & \secondx{0.393} & 0.424 & 0.427 & 0.426 & 0.402 & \secondx{0.422} & 0.413 & 0.431 & 0.438 & 0.450 & 0.440 & 0.462 & 0.458 & 0.450 & 0.440 & 0.460 & 0.423 & 0.437 & 0.407 & 0.424\\
\midrule

\multirow{5}{*}{ETTh2}
& 96  & \bestx{0.266} & \bestx{0.325} & 0.288 & 0.347 & 0.285 & 0.342 & 0.278 & 0.338 & \secondx{0.274} & \secondx{0.336} & 0.300 & 0.350 & 0.299 & 0.364 & 0.340 & 0.374 & 0.358 & 0.397 & 0.289 & 0.353 & \bestx{0.274} & 0.337\\
& 192 & \bestx{0.331} & \bestx{0.369} & 0.351 & 0.390 & 0.354 & 0.389 & 0.341 & 0.380 & 0.339 & 0.379 & 0.381 & 0.400 & 0.422 & 0.441 & 0.402 & 0.414 & 0.429 & 0.439 & 0.383 & 0.418 & \secondx{0.351} & 0.383\\
& 336 & \secondx{0.361} & \secondx{0.395} & 0.376 & 0.413 & 0.373 & 0.407 & 0.369 & 0.406 & \bestx{0.329} & \bestx{0.380} & 0.424 & 0.433 & 0.447 & 0.474 & 0.452 & 0.452 & 0.496 & 0.487 & 0.448 & 0.465 & 0.375 & 0.411\\
& 720 & \secondx{0.384} & \secondx{0.422} & 0.410 & 0.447 & 0.406 & 0.441 & 0.404 & 0.438 & \bestx{0.379} & \secondx{0.422} & 0.431 & 0.446 & 0.442 & 0.467 & 0.462 & 0.468 & 0.463 & 0.474 & 0.605 & 0.551 & 0.402 & 0.438\\
& avg & \secondx{0.336} & \secondx{0.378} & 0.356 & 0.399 & 0.354 & 0.394 & 0.348 & 0.391 & \second{0.330} & 0.379 & 0.367 & 0.407 & 0.402 & 0.437 & 0.414 & 0.427 & 0.437 & 0.449 & 0.431 & 0.447 & 0.351 & 0.392\\
\midrule

\multirow{5}{*}{Traffic}
& 96  & \bestx{0.345} & \bestx{0.233} & 0.359 & 0.254 & 0.388 & 0.282 & 0.371 & 0.258 & 0.360 & \secondx{0.249} & \secondx{0.351} & 0.257 & 0.519 & 0.309 & 0.593 & 0.321 & 0.587 & 0.366 & 0.410 & 0.282 & 0.366 & 0.259\\
& 192 & \bestx{0.364} & \bestx{0.242} & 0.380 & 0.264 & 0.407 & 0.290 & 0.388 & 0.264 & 0.379 & \secondx{0.256} & \secondx{0.365} & 0.265 & 0.537 & 0.315 & 0.617 & 0.336 & 0.604 & 0.373 & 0.423 & 0.287 & 0.381 & 0.265\\
& 336 & \bestx{0.379} & \bestx{0.249} & 0.398 & 0.273 & 0.412 & 0.294 & 0.393 & 0.266 & 0.392 & \secondx{0.264} & \secondx{0.382} & 0.273 & 0.534 & 0.313 & 0.629 & 0.336 & 0.621 & 0.383 & 0.436 & 0.296 & 0.397 & 0.269\\
& 720 & \bestx{0.419} & \bestx{0.271} & 0.438 & 0.296 & 0.450 & 0.312 & 0.427 & 0.324 & 0.432 & \secondx{0.286} & \secondx{0.421} & 0.292 & 0.577 & 0.325 & 0.640 & 0.350 & 0.626 & 0.382 & 0.466 & 0.315 & 0.429 & 0.292\\
& avg & \bestx{0.377} & \bestx{0.249} & 0.394 & 0.272 & 0.414 & 0.294 & 0.395 & 0.278 & 0.391 & \secondx{0.264} & \secondx{0.380} & 0.272 & 0.542 & 0.316 & 0.620 & 0.336 & 0.610 & 0.376 & 0.434 & 0.295 & 0.468 & 0.271\\
\midrule

\multirow{5}{*}{Electricity}
& 96  & \bestx{0.125} & \bestx{0.217} & 0.132 & 0.227 & 0.139 & 0.238 & 0.137 & 0.234 & \secondx{0.129} & \secondx{0.222} & 0.153 & 0.237 & 0.164 & 0.269 & 0.168 & 0.272 & 0.193 & 0.308 & 0.153 & 0.237 & 0.129 & 0.224\\
& 192 & \bestx{0.142} & \bestx{0.233} & 0.150 & 0.244 & 0.153 & 0.251 & \secondx{0.144} & 0.241 & 0.147 & \secondx{0.240} & 0.152 & 0.249 & 0.177 & 0.285 & 0.184 & 0.289 & 0.201 & 0.315 & 0.152 & 0.249 & 0.147 & 0.238\\
& 336 & \bestx{0.158} & \bestx{0.250} & 0.167 & 0.262 & 0.169 & 0.266 & \secondx{0.161} & 0.269 & 0.163 & \secondx{0.259} & 0.169 & 0.267 & 0.193 & 0.304 & 0.198 & 0.300 & 0.214 & 0.329 & 0.169 & 0.267 & \secondx{0.160} & 0.253\\
& 720 & \secondx{0.197} & \bestx{0.285} & 0.205 & 0.295 & 0.206 & 0.297 & 0.206 & \secondx{0.286} & \secondx{0.197} & 0.290 & 0.233 & 0.344 & 0.212 & 0.321 & 0.220 & 0.320 & 0.246 & 0.355 & 0.233 & 0.344 & \bestx{0.193} & 0.286\\
& avg & \bestx{0.156} & \bestx{0.246} & 0.164 & 0.257 & 0.167 & 0.263 & 0.162 & 0.258 & 0.159 & 0.253 & 0.177 & 0.274 & 0.187 & 0.295 & 0.192 & 0.295 & 0.214 & 0.327 & 0.177 & 0.224 & \secondx{0.157} & \secondx{0.250}\\
\midrule

\multirow{5}{*}{Weather}
& 96  & 0.148 & \bestx{0.189} & 0.155 & 0.204 & 0.162 & 0.212 & \secondx{0.147} & \secondx{0.195} & 0.149 & 0.198 & 0.163 & 0.211 & 0.161 & 0.229 & 0.172 & 0.220 & 0.217 & 0.296 & 0.152 & 0.237 & \bestx{0.145} & 0.197\\
& 192 & 0.193 & \bestx{0.234} & 0.203 & 0.248 & 0.204 & 0.248 & \secondx{0.192} & \secondx{0.237} & 0.194 & 0.241 & 0.205 & 0.250 & 0.220 & 0.281 & 0.219 & 0.261 & 0.276 & 0.336 & 0.220 & 0.282 & \bestx{0.187} & 0.238\\
& 336 & 0.244 & \bestx{0.274} & 0.259 & 0.290 & 0.254 & 0.286 & \secondx{0.242} & 0.281 & 0.245 & 0.282 & 0.245 & 0.289 & 0.278 & 0.331 & 0.280 & 0.306 & 0.339 & 0.380 & 0.265 & 0.319 & \bestx{0.240} & \secondx{0.280}\\
& 720 & \secondx{0.314} & \bestx{0.324} & 0.336 & 0.342 & 0.326 & 0.337 & \best{0.310} & 0.331 & 0.314 & 0.334 & 0.329 & 0.340 & 0.311 & 0.356 & 0.365 & 0.359 & 0.403 & 0.428 & 0.323 & 0.362 & 0.315 & 0.330\\
& avg & \secondx{0.225} & \bestx{0.255} & 0.238 & 0.271 & 0.237 & 0.270 & \secondx{0.223} & \secondx{0.261} & 0.226 & 0.264 & 0.238 & 0.273 & 0.243 & 0.299 & 0.259 & 0.287 & 0.309 & 0.360 & 0.240 & 0.300 & \bestx{0.222} & 0.261\\

\midrule
\multicolumn{2}{l}{\textbf{1st Count}}
& \multicolumn{2}{c}{51}
& \multicolumn{2}{c}{0}
& \multicolumn{2}{c}{0}
& \multicolumn{2}{c}{9}
& \multicolumn{2}{c}{3}
& \multicolumn{2}{c}{0}
& \multicolumn{2}{c}{0}
& \multicolumn{2}{c}{0}
& \multicolumn{2}{c}{0}
& \multicolumn{2}{c}{0}
& \multicolumn{2}{c}{7}
\\
\bottomrule
\end{tabular}
\end{adjustbox}
} 
\end{minipage}

\end{table}
\clearpage

\label{app5}
\section{Sensitivity Analysis of FAN Layer Depth}

We investigated the impact of FAN layer depth ($L$) within the Fourier Experts by comparing $L=1, 2, 3$ on the ETTh1 and ETTm1 datasets. The results in Table \ref{tab:fan_layers_ablation} show that the default single-layer design ($L=1$) consistently achieves the best performance. Increasing the depth to 2 or 3 layers yields no accuracy gains and leads to overfitting. Therefore, we adopt $L=1$ for efficiency and robustness.

\begin{table}[H]
\centering
\caption{Ablation study on the number of FAN layers ($L$) in the Fourier Experts. We compare $L=1, 2, 3$ on ETTh1 and ETTm1 datasets. \textbf{1-layer} represents the default setting in FM-LLM. The best results are highlighted in \textbf{bold}.}
\label{tab:fan_layers_ablation}
\resizebox{\textwidth}{!}{%
\begin{tabular}{c|c|cc|cc|cc|cc}
\toprule
\multirow{2}{*}{\textbf{Dataset}} & \multirow{2}{*}{\textbf{FAN Layers}} & \multicolumn{2}{c|}{\textbf{96}} & \multicolumn{2}{c|}{\textbf{192}} & \multicolumn{2}{c|}{\textbf{336}} & \multicolumn{2}{c}{\textbf{720}} \\ \cmidrule(l){3-10} 
 &  & MSE & MAE & MSE & MAE & MSE & MAE & MSE & MAE \\ \midrule
\multirow{3}{*}{ETTh1} & 1-layer (Ours) & \textbf{0.342} & \textbf{0.380} & \textbf{0.377} & \textbf{0.403} & \textbf{0.395} & \textbf{0.415} & \textbf{0.397} & \textbf{0.429} \\
 & 2-layer & 0.343 & 0.382 & 0.378 & 0.405 & 0.397 & 0.419 & 0.411 & 0.441 \\
 & 3-layer & 0.378 & 0.406 & 0.421 & 0.433 & 0.448 & 0.451 & 0.471 & 0.480 \\ \midrule
\multirow{3}{*}{ETTm1} & 1-layer (Ours) & \textbf{0.276} & \textbf{0.327} & \textbf{0.319} & \textbf{0.356} & \textbf{0.351} & \textbf{0.379} & \textbf{0.406} & \textbf{0.415} \\
 & 2-layer & 0.289 & 0.339 & 0.334 & 0.369 & 0.368 & 0.390 & 0.428 & 0.423 \\
 & 3-layer & 0.280 & 0.333 & 0.325 & 0.361 & 0.361 & 0.384 & 0.432 & 0.423 \\ \bottomrule
\end{tabular}%
}
\end{table}

\label{app6}
\section{Comparison with Statistical Baseline}
We employed the pmdarima library for implementation. For each variable in the multivariate time series, a univariate ARIMA~\cite{box1970distribution} model was fitted independently. We used the \texttt{auto\_arima} algorithm to search for optimal hyperparameters with the following constraints: maximum order of $p$ and $q$ set to 3 to prevent overfitting, and a seasonal period $m=24$ to capture daily cycles. Consistent with the few-shot setting of FM-LLM, the ARIMA models were trained strictly on the initial 10\% of the data. During inference, we utilized a rolling forecast strategy with a fixed model (no re-training) to evaluate generalization capability.

The comparative results on ETTh1, ETTh2, ETTm1, and ETTm2 under the 10\% few-shot setting are summarized in Table \ref{tab:arima_comparison}.
It is observed that ARIMA struggles significantly in long-horizon forecasting (e.g., $H=336, 720$), exhibiting high MSE values. This is primarily due to the error accumulation inherent in recursive steps for linear models and their inability to model complex multivariate correlations. In contrast, FM-LLM maintains robust performance, demonstrating its superiority in capturing long-term dependencies and utilizing pre-trained knowledge for generalization.

\begin{table}[H]
\centering
\caption{Few-shot performance comparison between ARIMA and FM-LLM on ETT datasets (10\% training data).The best results are highlighted in \textbf{bold}.}
\label{tab:arima_comparison}
\resizebox{\textwidth}{!}{%
\begin{tabular}{c|c|cc|cc|cc|cc}
\toprule
\multirow{2}{*}{\textbf{Dataset}} & \multirow{2}{*}{\textbf{Model}} & \multicolumn{2}{c|}{\textbf{96}} & \multicolumn{2}{c|}{\textbf{192}} & \multicolumn{2}{c|}{\textbf{336}} & \multicolumn{2}{c}{\textbf{720}} \\ \cmidrule(l){3-10} 
 &  & MSE & MAE & MSE & MAE & MSE & MAE & MSE & MAE \\ \midrule
\multirow{2}{*}{ETTh1} & ARIMA & 1.496 & 0.700 & 1.462 & 0.745 & 2.361 & 1.000 & 4.773 & 1.357 \\
 & \textbf{FM-LLM} & \textbf{0.451} & \textbf{0.454} & \textbf{0.481} & \textbf{0.472} & \textbf{0.497} & \textbf{0.486} & \textbf{0.525} & \textbf{0.514} \\ \midrule
\multirow{2}{*}{ETTh2} & ARIMA & 1.512 & 0.780 & 1.847 & 0.861 & 2.172 & 0.922 & 2.541 & 1.020 \\
 & \textbf{FM-LLM} & \textbf{0.298} & \textbf{0.362} & \textbf{0.340} & \textbf{0.391} & \textbf{0.354} & \textbf{0.404} & \textbf{0.398} & \textbf{0.438} \\ \midrule
\multirow{2}{*}{ETTm1} & ARIMA & 2.266 & 0.974 & 2.603 & 1.051 & 2.874 & 1.117 & 3.051 & 1.172 \\
 & \textbf{FM-LLM} & \textbf{0.383} & \textbf{0.403} & \textbf{0.422} & \textbf{0.427} & \textbf{0.463} & \textbf{0.450} & \textbf{0.561} & \textbf{0.498} \\ \midrule
\multirow{2}{*}{ETTm2} & ARIMA & 1.385 & 0.784 & 1.559 & 0.835 & 1.769 & 0.892 & 2.136 & 0.976 \\
 & \textbf{FM-LLM} & \textbf{0.172} & \textbf{0.261} & \textbf{0.224} & \textbf{0.295} & \textbf{0.273} & \textbf{0.327} & \textbf{0.366} & \textbf{0.382} \\ \bottomrule
\end{tabular}%
}
\end{table}

\label{app7}

\section{Diagnosing the Non-monotonic Effect of Training Data Size}
\begin{figure}[!t]
    \centering

    \begin{subfigure}[b]{1\linewidth}
        \centering
        \includegraphics[width=\linewidth]{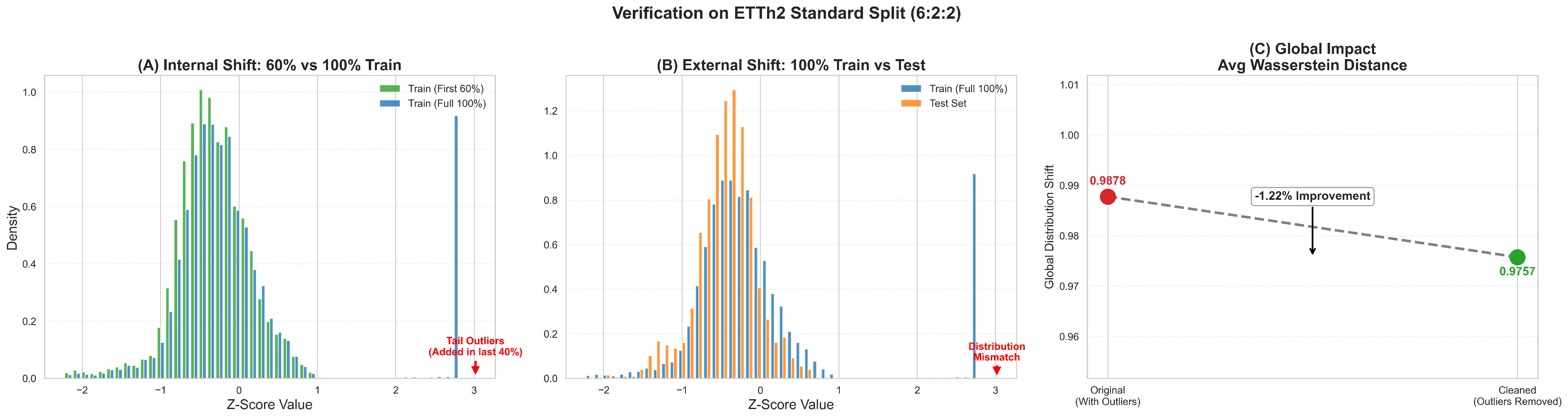}
        \caption{ETTh2 (6:2:2 split): internal train shift (first 60\% vs 100\%), train–test mismatch, and the average Wasserstein distance across variables before/after removing MUFL-outlier time indices.}
        \label{fig:app7_a}
    \end{subfigure}

    \vspace{0.6em} %

    \begin{subfigure}[b]{1\linewidth}
        \centering
        \includegraphics[width=\linewidth]{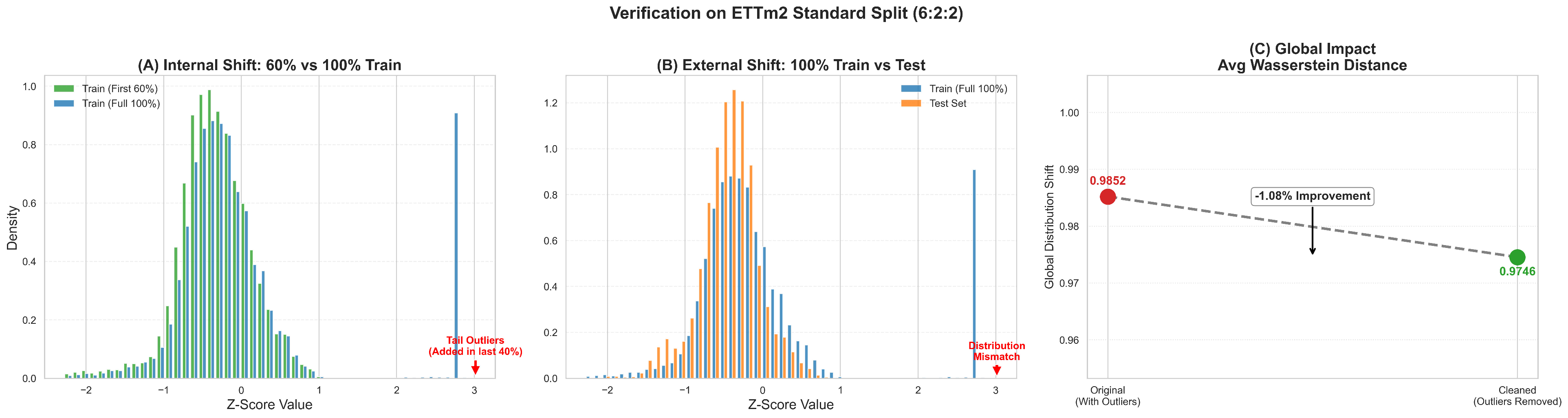}
        \caption{ETTm2 (6:2:2 split): same diagnosis as (a).}
        \label{fig:app7_b}
    \end{subfigure}

    \caption{Distribution-shift diagnosis for the non-monotonic data-scale trend on ETTh2/ETTm2. Panels (A)–(B) visualize internal and external distribution mismatch on MUFL (z-score), and panel (C) reports the global impact via the average Wasserstein distance across variables before/after removing MUFL-outlier time indices.}
    \label{fig:app7}
\end{figure}

To further understand the counter-intuitive non-monotonic data-scale trend, we provide an auxiliary distribution-shift diagnosis on ETTh2 and ETTm2 under the standard 6:2:2 split.
As shown in Fig.~\ref{fig:app7}, we first visualize the internal shift within the training split by comparing the first 60\% of training samples against the full training split on the MUFL channel (z-score), where tail outliers mainly appear in the last 40\% (Fig.~\ref{fig:app7}a).
We then visualize the external mismatch between the full training split and the test split on MUFL (Fig.~\ref{fig:app7}b).
Finally, we assess the global impact of these atypical tail segments by removing the time indices flagged as MUFL outliers and recomputing the average Wasserstein distance between training and test across all variables (Fig.~\ref{fig:app7}c).
Overall, this diagnostic analysis suggests that atypical tail segments can affect the measured train--test distribution mismatch, which may contribute to the observed non-monotonic behavior when increasing the training data size.

\clearpage

\bibliographystyle{elsarticle-num}
\bibliography{refs}

\end{document}